\documentclass[review]{elsarticle}
\usepackage{float}
\usepackage{lineno,hyperref}
\usepackage[margin=1in]{geometry} 
\usepackage{setspace}
\usepackage{mathtools}
\usepackage{tablefootnote, url}
\usepackage{afterpage}
\usepackage{floatpag}
\usepackage{amsmath,amssymb,amsfonts}

\usepackage{marginnote,zref-savepos}
\newcounter{labelnote}
\makeatletter
\let\oldmarginnote\marginnote
\renewcommand*{\marginnote}[1]{%
 \begingroup\strut
  \stepcounter{labelnote}\zsaveposx {marginnote-\thelabelnote}
     \ifnum 0\zposx{marginnote-\thelabelnote}<1900000
      \reversemarginpar
      \oldmarginnote{\color{blue}#1}%
     \else
      \normalmarginpar
      \oldmarginnote{\color{blue}#1}%
     \fi
 \endgroup%
}
\makeatother
\usepackage{graphicx}
\usepackage{xcolor}
\modulolinenumbers[5]
\journal{Journal of Artificial Intelligence}

\begin{document}

\begin{frontmatter}

\title{Risk-averse autonomous systems:\\A brief history and recent developments from the perspective of\\optimal control}
\tnotetext[mytitlenote]{Submitted in August 2021; revised in February and May 2022; accepted in May 2022. This work is part of the Special Issue on Risk-aware Autonomous Systems: Theory and Practice in the \emph{Journal of Artificial Intelligence}.}
\author{Yuheng Wang\fnref{affiliation} and Margaret P. Chapman\fnref{affiliation}}
\address{Edward S. Rogers Sr. Department of Electrical and Computer Engineering, University of Toronto, 10 King's College Road, Toronto, Ontario, M5S 3G8, Canada}
\fntext[affiliation]{Y.W. and M.P.C. are with the Edward S. Rogers Sr. Department of Electrical and Computer Engineering, University of Toronto. \ead{danielwang.wang@mail.utoronto.ca, mchapman@ece.utoronto.ca}}


\begin{singlespace}
\begin{abstract}
We present an historical overview about the connections between the analysis of risk and the control of autonomous systems. We offer two main contributions. Our first contribution is to propose three overlapping paradigms to classify the vast body of literature: the \emph{worst-case}, \emph{risk-neutral}, and \emph{risk-averse paradigms}. We consider an appropriate assessment for the risk of an autonomous system to depend on the application at hand. In contrast, it is typical to assess risk using an expectation, variance, or probability alone. Our second contribution is to unify the concepts of risk and autonomous systems. We achieve this by connecting approaches for quantifying and optimizing the risk that arises from a system's behaviour across academic fields. The survey is highly multidisciplinary. We include research from the communities of reinforcement learning, stochastic and robust control theory, operations research, and formal verification. We describe both model-based and model-free methods, with emphasis on the former. Lastly, we highlight fruitful areas for further research. A key direction is to blend risk-averse model-based and model-free methods to enhance the real-time adaptive capabilities of systems to improve human and environmental welfare.
\end{abstract}

\end{singlespace}
\end{frontmatter}


\begin{singlespace}
\section{Introduction}
{In the English language, the notion of risk has different definitions, which include the ``possibility of loss or injury'' and an entity that ``creates or suggests a hazard'' \cite{miriamwebster}. The International Organization for Standardization (ISO), a global federation of national standards bodies, defines risk as the ``effect of uncertainty on objectives,'' which may be interpreted qualitatively or quantitatively based on the application at hand \cite{isosource}. In this survey, we study the connections between risk and control of autonomous systems. We adopt the ISO definition for risk and emphasize a quantitative viewpoint, in which performance criteria for a system of interest have been specified.} 


Historically, two main paradigms have been developed for quantifying and managing the consequences that may arise from a system's behaviour: the risk-neutral paradigm and the worst-case (i.e., robust) paradigm. The terms \emph{risk-neutral} and \emph{worst-case} describe different ways in which uncertain outcomes can be summarized. An uncertain outcome is a cost that may arise due to a system's behaviour. 
A cost can be a stochastic quantity, such as a random variable, or a nonstochastic quantity, such as a real number. A cost distribution is a probability distribution of a cost. If the cost is nonstochastic, then its distribution is a point mass; that is, a single value occurs with certainty. The \emph{risk-neutral paradigm} summarizes uncertain outcomes in terms of an average outcome, which is defined using an expectation. In contrast, the \emph{worst-case paradigm} summarizes uncertain outcomes in terms of the most harmful outcome. The meaning of ``most harmful'' depends on prespecified constraints on the values of uncertain uncontrollable inputs, which are called disturbances. 

A typical optimal control problem in these paradigms takes the following form:
\begin{equation}\label{standard1}\begin{aligned}
& \underset{\pi \in \Pi}{\text{minimize}}\;\;\;\;\;\; \;\;J_{x}^{\pi}(Z) \\
& \text{subject to } \;\;\;\;\; \psi_{x,i}^{\pi}(Z_i) \in b_i, \quad i \in \mathcal{I},\\
& \hphantom{\text{subject to }}\;\;\;\;\;\text{system dynamics under } \pi,
\end{aligned}\end{equation}
where $x$ is an initial state, $\pi$ is a control policy, $\Pi$ is a class of control policies, $b_i$ is an interval, $Z$ and $Z_i$ are costs, and $\mathcal{I}$ is a collection of indices. $J_{x}^\pi$ and $\psi_{x,i}^\pi$ are maps from a space of costs to the extended real line, which quantify an objective criterion and a constraint, respectively. For example, $\psi_{x,i}^\pi(Z_i)$ may represent the maximum water level of a reservoir at time $i$, which should remain within a given interval $b_i = [0,h]$ to avoid an overflow.\footnote{The water level of a reservoir depends on disturbances, such as wind or precipitation.} The (system) dynamics are equations that describe the behaviour of a system over time. These equations can be known, unknown, or partially known. The dynamics of a \emph{stochastic} system involve random processes (e.g., sequences of random variables). In contrast, the dynamics of a \emph{nonstochastic} system only involve the evolution of deterministic quantities (e.g., vectors in Euclidean spaces). The time horizon can be discrete, continuous, or a combination. We reserve a formal description of a discrete-time stochastic system for Section \ref{systemmodelsec}. The term policy is short-hand for the term control policy. An \emph{optimal policy} is a policy that minimizes the value of the objective while satisfying all constraints; such a policy need not exist. The meaning of the term \emph{optimal} depends on the problem at hand, and we will see that there are many ways to define this term.

A standard problem in the risk-neutral paradigm is to find a policy that minimizes an expected cumulative cost. In this problem, $J_{x}^\pi$ is an expectation, $Z$ is a sum of random variables, and $\mathcal{I}$ is empty:
\begin{equation}\label{standard2}\begin{aligned}
 & \underset{\pi \in \Pi}{\text{minimize}}\;\;\;\;\;\; \;\;E_{x}^\pi(Z) \\
 & \text{subject to }\;\;\;\;\;\text{system dynamics under } \pi,
\end{aligned}\end{equation}
where the distribution of $Z$ depends on an initial state $x$, a policy $\pi$, and the system dynamics. 
Comprehensive presentations of this problem 
from the reinforcement learning and stochastic control communities are provided by \cite{sutton2018reinforcement} and \cite{bertsekas2004stochastic}, respectively. 

\begin{figure}[!h]
\centerline{\includegraphics[width=0.7\textwidth]{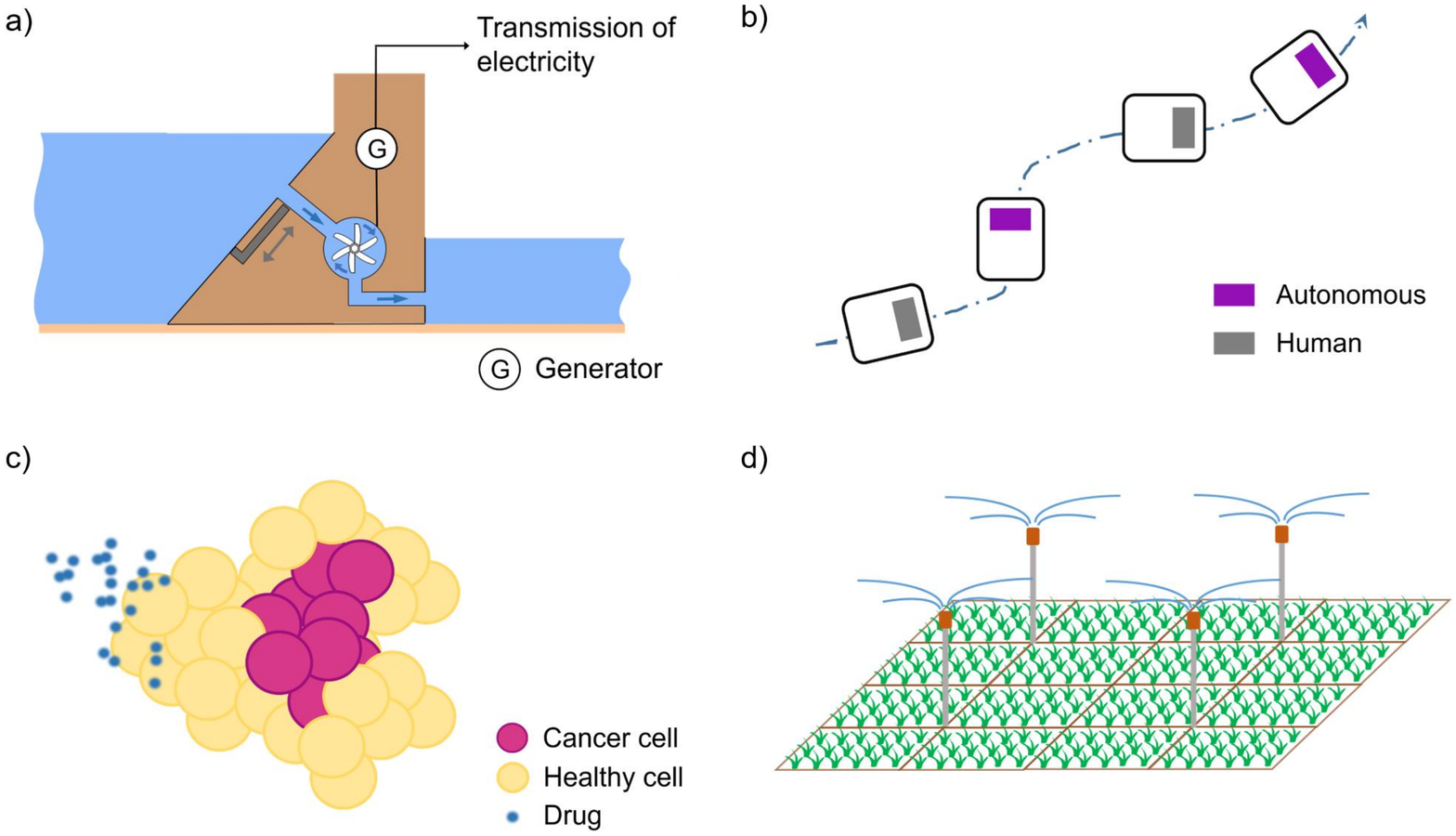}}
\vspace{-8mm}
\caption{Examples of systems from different disciplines with safety concerns. a) Stormwater systems are required to minimize overflows while meeting other goals, such as storing potential energy for future electricity generation, providing sufficient irrigation to vegetation in urban areas, or retaining water long enough to improve its quality. b) Networks of autonomous and human-driven vehicles must avoid collisions, but if collisions are unavoidable, then accelerations that humans experience should be minimized to reduce the severity of injuries. c) Cancer treatments must manage the growth of cancer cells without posing too much harm to healthy cells. d) Agriculture systems must generate sufficient crop yields while conserving water and energy. For instance, energy may be required to pump water from reservoirs.}
\label{examplesofsystemswithsafetygoals}
\end{figure}

A key limitation of the risk-neutral paradigm is its ignorance of the characteristics of a cost distribution other than the mean. For example, we may prefer cost distributions with smaller variance, smaller mean in the upper tail, or smaller upper-semideviation $E(\max\{Z - E(Z),0\})$. Such preferences may be important for systems with safety concerns. While the meaning of safety is application-dependent, a common safety specification for an autonomous system is to avoid operating in a particular region. Moreover, a system may be required to operate in a manner that alleviates harmful consequences, even in rare situations that are difficult to predict. Systems with safety concerns are ubiquitous. Stormwater infrastructure must minimize overflows while meeting other goals, such as storing potential energy for future electricity needs; networks of human-driven and autonomous vehicles must avoid collisions; and cancer treatments must manage the growth of cancer while minimizing adverse side effects (Figure \ref{examplesofsystemswithsafetygoals}). 
\begin{singlespace}
\end{singlespace}

Traditionally, one computes policies for autonomous systems with safety concerns by formulating a problem in the worst-case paradigm. A standard problem in this paradigm is to find a policy that minimizes a maximum cost \cite{bertsekas1971minimax, heger1994consideration, coraluppi1999risk, morimoto2005robust, margellos2011hamilton, chen2018hamilton}. The maximum cost can quantify how bounded adversarial disturbances can inhibit the satisfactory operation of a system. When $Z$ is nonstochastic, a common form of the objective criterion $J_{x_0}^\pi(Z)$ \eqref{standard1} is
\begin{equation}\label{my3}
 J_{x_0}^\pi(Z) = \sup_{w_0 \in F_0, \dots, w_{N-1} \in F_{N-1}} \left( c_N(x_N) + \sum_{t=0}^{N-1} c_t(x_t,\pi_t(x_t),w_t) \right),
\end{equation}
where $F_t$ is a bounded subset of $\mathbb{R}^d$ for every $t$, representing the set of possible disturbances at time $t$. Here, a disturbance $w_t$ is viewed as an adversary because its value is chosen to maximize the cost on the right side of \eqref{my3}. Let us further describe the notation in \eqref{my3}: $x_t$ is a state; $\pi_t(x_t)$ is an action; $\pi_t$ is a map from states to actions, which may be time-dependent; $\pi = (\pi_0,\pi_1,\dots,\pi_{N-1})$ is called a policy; $c_t$ is a stage or terminal cost function; and $N$ is a natural number that specifies the length of a finite discrete-time horizon. The worst-case paradigm is suitable when 1) it is possible to characterize the bounds of disturbances with a sufficient degree of certainty, 2) such bounds do not change in unanticipated ways over time, and 3) a policy can be computed that is not too cautious to apply in practice. In particular, this paradigm has been applied to the aerospace domain with tremendous success. However, a fundamental issue with the worst-case paradigm is the typical assumption of bounded adversarial disturbances, which may not be suitable for every application.

The risk-neutral and worst-case paradigms represent two distinct perspectives regarding the future, that is, a future of averages and a future of misfortune, respectively. An intermediary paradigm, called the \emph{risk-averse paradigm}, forms a bridge between the risk-neutral and worst-case paradigms. {Colloquially, the term risk-averse describes people or algorithms that prefer outcomes with reduced uncertainty. Here, we propose a definition that emphasizes autonomous systems and their cost distributions. A cost distribution of an autonomous system describes the potential consequences that may arise from the system's behaviour. The consequences are defined based on the meaning of satisfactory operation, which is application-specific. In this survey, the term \emph{risk-averse} describes people or algorithms that prefer cost distributions with particular characteristics, where the characteristics reflect a desire to reduce harm. \emph{To quantify risk} means to summarize numerically the consequences that may arise from a system's behaviour.}

The notions of risk, autonomous systems, and control are connected through the formulation of a \emph{risk-averse optimal control problem}. Such a problem is designed to produce a cost distribution for an autonomous system with specific desirable attributes. For example, a problem may encode a preference for cost distributions with a reduced 
\begin{enumerate}
    \item linear combination of mean and variance,\label{ex1}
    \item average cost subject to a probabilistic constraint,
    \item expected utility, where the utility function reflects personal subjective preferences,
    \item quantile, or
    \item average cost above a quantile.\label{exend}
\end{enumerate}
A risk-averse optimal control problem takes the form of \eqref{standard1}, where the maps $J_{x}^{\pi}$ and $\psi_{x,i}^{\pi}$ encode preferences for cost distributions with particular characteristics. 
For instance, these maps may be defined in terms of the items \ref{ex1}--\ref{exend} above, an expected cost, or a maximum cost.
We use the term \emph{risk-averse approach} (or risk-averse method) to describe an algorithm that solves a risk-averse optimal control problem. \emph{To solve} means to return the minimum value of the objective $J_{x}^{\pi}(Z)$
exactly or approximately and may include returning an optimal policy under appropriate conditions. 

In this survey, we provide an historical overview of research in the intersection of risk analysis and autonomous systems. This research area is important and timely. Pressing challenges related to human and environmental health are characterized by systems that evolve in uncertain ways (e.g., a patient, a disease infecting a population, an electrical grid with renewable energy sources, a supply chain for a computer, an urban water network, and a population's production of waste). Advancing both theoretical and applied research about risk analysis and autonomous systems may lead to improved solution pathways for critical modern challenges. 
Moreover, a vast body of literature in this intersection has emerged over the past fifty years from various disciplines. We provide a multidisciplinary overview to guide and inspire future studies. First, we propose three paradigms to classify the literature: the worst-case, risk-neutral, and risk-averse paradigms. While it is standard to assess risk using an expectation, variance, or probability, we view an appropriate assessment to depend on the application of interest. Second, we formalize the connection between risk and autonomous systems through risk-averse optimal control problems. We describe different ways to quantify risk using mathematical tools called \emph{risk functionals}, and we present both model-based and model-free algorithms, with emphasis on the former. Third, we discuss future research directions about expanding real-time adaptive capabilities and application areas.

We wish to bring the reader's attention to related topics that we do not emphasize, as existing reviews are available. In 2014, Pecka and Svoboda \cite{pecka2014safe} summarized safe exploration approaches in reinforcement learning (RL) using the classification from \cite{garcia2012safe}, which includes learning from demonstrations \cite{ravichandar2020recent}. In 2015, Garc{\'i}a and Fern{\'a}ndez surveyed RL algorithms that consider safety in policy synthesis, with emphasis on model-free methods \cite[Table 2]{garcia2015comprehensive}.
In 2020, Hewing et al. reviewed approaches that combine learning with model predictive control and highlighted safety issues \cite{hewing2020learning}. A survey from 2022 describes safe learning approaches for control, with focus on robotics applications \cite{Brunke}. Recent surveys on inverse reinforcement learning are provided by \cite{ab2020inverse, arora2021survey}. 

In contrast, our survey presents approaches for quantifying risk (Section \ref{howtoquantifyrisk}) and solving risk-averse optimal control problems, with emphasis on risk functionals and a model-based viewpoint (Section \ref{history}). As well as contributions from the RL community, we present related contributions from the stochastic control, operations research, and formal verification communities. We overview risk-averse model-free methods in Section \ref{emerging}, and we discuss future directions towards enhancing the real-time adaptive capabilities of risk-averse autonomous systems in Section \ref{conclusion}. 

\paragraph{Relevance to artificial intelligence} The broad problem of how to analyze and optimize the behaviour of an autonomous system is highly relevant to the field of artificial intelligence. For example, designing algorithms to ensure that robots can safely and efficiently traverse uneven terrain is an instance of this problem. Adapting treatment protocols to a particular cancer patient based on their oncologist's expertise, data from their prior treatment cycles, and biochemical chemotherapy models is another instance of this problem. Our survey offers an historical and modern presentation of research in the intersection of \emph{risk analysis} and \emph{autonomous systems}. This intersection is quite multifaceted and incorporates ideas from several domains, including reinforcement learning, decision and control theory, and operations research. We are hopeful that this survey about risk-averse autonomous systems will inspire future theoretical and applied research that will enhance the operation of systems in practice and therefore the quality of life.

\paragraph{Organization} Section \ref{systemmodelsec} presents a Markov decision process model that we use throughout the survey. Section \ref{howtoquantifyrisk} presents approaches for quantifying risk using maps on spaces of random variables (such maps are called risk functionals). Section \ref{history} describes methods for risk-averse optimal control, with emphasis on model-based methods. Section \ref{emerging} presents model-free methods for risk-averse optimal control. Section \ref{conclusion} provides concluding remarks and future directions.

\paragraph{Notation and terminology} Throughout the survey, we provide mathematical details in footnotes for the interested reader, which can be skipped. $\mathbb{R}$ is the real line. $\mathbb{R}_+ \coloneqq \{ y \in \mathbb{R} : y \geq 0 \}$ is the nonnegative orthant of $\mathbb{R}$. $\mathbb{N} \coloneqq \{1,2,3,\dots\}$ is the set of natural numbers. Given $N \in \mathbb{N}$, we define $\mathbb{T} \coloneqq \{0,1,\dots,N-1\}$. A probability space is a tuple $(\Omega,\mathcal{F},\mu)$, where $\Omega$ is a nonempty set, $\mathcal{F}$ is a \emph{sigma algebra} on $\Omega$, and $\mu$ is a probability measure on $\mathcal{F}$. $\mathcal{F}$ is a nonempty collection of subsets of $\Omega$ that is closed under countable unions and complements. A member of $\mathcal{F}$ is called an event; a synonym is a measurable subset of $\Omega$. If $\mathcal{M}$ is a metric space, then $\mathcal{B}_\mathcal{M}$ denotes the \emph{Borel sigma algebra} on $\mathcal{M}$. $\mathcal{B}_\mathcal{M}$ is the smallest sigma algebra on $\mathcal{M}$ that contains all open subsets of $\mathcal{M}$. The notation $B \in \mathcal{B}_\mathcal{M}$ means that $B$ is a member of $\mathcal{B}_\mathcal{M}$; $B$ is called a \emph{Borel-measurable subset} of $\mathcal{M}$. A \emph{random cost} is a random variable for which smaller realizations are preferred.\footnote{A random variable is a type of measurable function. Informally, a measurable function is a function that is ``regular enough'' to be integrated. The concepts of measurable spaces and Borel sigma algebras characterize random variables. In this footnote, we outline these concepts. We restrict our attention to Borel sigma algebras on metric spaces rather than topological spaces for simplicity. A \emph{measurable space} $(\Omega, \mathcal{F})$ is a pair that consists of a nonempty set $\Omega$ and a sigma algebra $\mathcal{F}$ of subsets of $\Omega$. Let $(\Omega,\mathcal{F})$ and $(\Omega',\mathcal{F}')$ be measurable spaces, and let $f : \Omega \rightarrow \Omega'$ be a function. $f$ is $(\mathcal{F},\mathcal{F}')$-\emph{measurable} if and only if for every $B \in \mathcal{F}'$, the set $\{\omega \in \Omega : f(\omega) \in B\}$ is a member of $\mathcal{F}$. There are different ways to describe a measurable function $f$, and we present two examples below.
\begin{itemize}
    \item If $(\Omega',\mathcal{F}') = (\mathbb{R},\mathcal{B}_{\mathbb{R}})$, then $f$ is called a \emph{random variable}.
    \item If $\Omega$ and $\Omega'$ are metric spaces, $\mathcal{F} = \mathcal{B}_{\Omega}$, and $\mathcal{F}' = \mathcal{B}_{\Omega'}$, then $f$ is called a \emph{Borel-measurable function} (or is described as being Borel-measurable). \end{itemize} More details can be found in classical texts, such as \cite{bertsekas2004stochastic, folland1999real, ash1972probability}. \label{technicalfootnote}}  The notation $Z \in L^\text{p}(\Omega,\mathcal{F},\mu)$ with $\text{p} \in [1,+\infty)$ means that $Z : \Omega \rightarrow \mathbb{R}$ is a random variable whose $\text{p}$\textsuperscript{th} moment with respect to the probability measure $\mu$ is finite, i.e., 
\begin{equation}
    \| Z \|_\text{p} \coloneqq \big(E(|Z|^\text{p})\big)^{1/\text{p}} \coloneqq \left(\int_{\Omega} |Z|^\text{p} \; \mathrm{d}\mu\right)^{1/\text{p}} < +\infty.
\end{equation}
The symbol $\| \cdot \|_\text{p}$ denotes the $L^\text{p}$-norm, and the symbol $E(\cdot)$ denotes expectation (i.e., integration with respect to $\mu$). It is common to abbreviate $L^\text{p}(\Omega,\mathcal{F},\mu)$ by $L^\text{p}$ when the definition of the underlying probability space is not important. Let $Z : \Omega \rightarrow \mathbb{R}$ be a random variable. For every $\omega \in \Omega$, the number $Z(\omega) \in \mathbb{R}$ is called a \emph{realization} of $Z$.
The symbol $\sigma_Z^2$ denotes the variance of $Z$, and $\sigma_Z$ denotes the standard deviation of $Z$. The symbol $\inf$ means \emph{infimum}, which is the technical term for the smallest value. The symbol $\sup$ means \emph{supremum}, which is the technical term for the largest value. We write $\inf$ instead of $\min$ because the minimum may not exist; for example, the minimum of the set $\left\{\frac{1}{n} : n \in \mathbb{N}\right\}$ does not exist, but its infimum exists and is zero. Analogously, we write $\sup$ instead of $\max$. We assume that all functions are Borel-measurable.\footnote{Here, we explain why Borel-measurable functions are important (see Footnote \ref{technicalfootnote} for the definition). An autonomous system need not have finitely many states and actions. An algorithm for such a system requires functions to be Borel-measurable so that integrals, such as expectations, of these functions can be defined. While continuous functions on metric spaces are Borel-measurable \cite[Cor. 2.2]{folland1999real}, it is not always appropriate to assume continuous functions. For instance, bang-bang controllers are not continuous.} 
The terms \emph{theoretical guarantee} and \emph{in principle} describe a mathematical statement or property that holds under specific assumptions.

\section{System model}\label{systemmodelsec}
While an appropriate model is application-dependent, we consider a discrete-time stochastic system to elucidate the differences between methods. The model takes the form
\begin{subequations}\label{sysmodel}
\begin{equation}
    x_{t+1} = f_t(x_t,u_t,w_t), \quad \quad t = 0,1,2,\dots,
\end{equation}
where $x_t \in S$ is a state, $u_t \in C$ is an action, and $w_t \in D$ is a disturbance. The state space $S \in \mathcal{B}_{\mathbb{R}^n}$, action space $C \in \mathcal{B}_{\mathbb{R}^m}$, and disturbance space $D \in \mathcal{B}_{\mathbb{R}^d}$ are Borel-measurable subsets of Euclidean spaces. 
%
Typically, an initial state is given. $f_t$ is a function that describes how a current state $x_t$, action $u_t$, and disturbance $w_t$ lead to a future state $x_{t+1}$. The sequence $(w_0,w_1,w_2,\dots) \subseteq D$ is a realization of a random process $(W_0,W_1,W_2,\dots)$, where $W_t$ is independent of $W_\tau$ for every $\tau \neq t$. 
The distribution of $W_t$ is $p_t$, which is a probability measure on $\mathcal{B}_D$ and may not be known.\footnote{For convenience, define $\mathbb{N}_0 \coloneqq \mathbb{N} \cup \{0\}$. The $D$-valued process $(W_0,W_1,\dots)$ is defined on an arbitrary probability space $(\Omega,\mathcal{F},\mu)$, and for every $t \in \mathbb{N}_0$, the distribution of $W_t$ is $p_t$. These statements mean the following: for every $t \in \mathbb{N}_0$, the function $W_t : \Omega \rightarrow D$ is $(\mathcal{F},\mathcal{B}_D)$-measurable and $$p_t(B) = \mu(\{\omega \in \Omega : W_t(\omega) \in B \})$$ for every $B \in \mathcal{B}_D$ \cite[p. 220]{ash1972probability}. The process $(W_0,W_1,\dots)$ being independent means that for every finite set $\{i_1,\dots,i_{M}\}$ of distinct indices in $\mathbb{N}_0$, it holds that $$\mu(\{\omega \in \Omega : W_{i_1}(\omega) \in B_1, \dots, W_{i_{M}}(\omega) \in B_M\}) = \mu(\{\omega \in \Omega : W_{i_1}(\omega) \in B_1\}) \cdots \mu(\{\omega \in \Omega : W_{i_M}(\omega) \in B_M\})$$ for every $B_1\in \mathcal{B}_D, \dots, B_M \in \mathcal{B}_D$ \cite[pp. 213--214]{ash1972probability}.} $X_t : \mathbf{\Omega} \rightarrow S$ and $U_t : \mathbf{\Omega} \rightarrow C$ are functions, where $\mathbf{\Omega}$ is a space that contains all possible trajectories.\footnote{$\mathbf{\Omega}$ is an example of a sample space. The definition of $\mathbf{\Omega}$ depends on the definition of a trajectory, which is problem-specific. If the problem is to minimize an expected cumulative cost on an infinite discrete-time horizon, then it is typical to define $\mathbf{\Omega}$ by $\mathbf{\Omega} \coloneqq (S \times C)^\infty$. In this case, a trajectory $\omega \in \mathbf{\Omega}$ takes the form $\omega = (x_0,u_0,x_1,u_1,\dots)$, where $x_t \in S$ and $u_t \in C$ are the state and action at time $t$ in the trajectory $\omega$, respectively. For other problems, it may be useful to define a random additional state $Y_t$, whose realizations are in a space $\mathcal{Y}$ (see Section \ref{statespaceaug} for an example). In this latter case, one may define $\mathbf{\Omega}$ by $\mathbf{\Omega} \coloneqq (S \times \mathcal{Y} \times C)^\infty$. Here, a trajectory $\omega \in \mathbf{\Omega}$ takes the form $\omega = (x_0,y_0,u_0,x_1,y_1,u_1,\dots)$, where $(x_t,y_t) \in S \times \mathcal{Y}$ and $u_t \in C$ are the augmented state and action at time $t$ in the trajectory $\omega$, respectively. In any case, the coordinates of every $\omega \in \mathbf{\Omega}$ are related casually through equations such as \eqref{sysmodel} and the chosen policy class.} For every trajectory $\omega \in \mathbf{\Omega}$, one defines $X_t(\omega) \coloneqq x_t$ and $U_t(\omega) \coloneqq u_t$, where $x_t \in S$ and $u_t \in C$ are the state and action at time $t$ in the trajectory $\omega$, respectively. Hence, we call $X_t$ the \emph{random state} at time $t$, and we call $U_t$ the \emph{random action} at time $t$. The quantity $q_t(\underline{S}|x,u)$ is the probability that $X_{t+1}$ is realized in $\underline{S} \in \mathcal{B}_S$, provided that the realization of $(X_t,U_t)$ is $(x,u) \in S \times C$. This conditional probability is defined in terms of $p_t$ and $f_t$ as follows:
\begin{equation}\label{qt}
    q_t(\underline{S}|x,u) \coloneqq p_t(\{w \in D : f_t(x,u,w) \in \underline{S} \}).
\end{equation}
\end{subequations}
$q_t$ may be fully known, partially known, or unknown. In some settings, $q_t$ is only available indirectly by sampling from a simulator.

There is a class $\Pi$ of history-dependent policies that is associated with the system \eqref{sysmodel}. The precise definition of $\Pi$ is problem-specific. Section \ref{history} emphasizes problems with deterministic history-dependent policies and a finite discrete-time horizon $\{0,1,\dots,N\}$, where $N \in \mathbb{N}$ is given. In this case, a policy $\pi \in \Pi$ takes the form $\pi = (\pi_0,\pi_1, \dots,\pi_{N-1})$, where each $\pi_t$ is a function from prior states and actions to an action for time $t$. If $\pi_t$ depends on the entire history of prior states and actions, then it has the form
\begin{equation}
    \pi_t(x_0,u_0,\dots,x_{t-1},u_{t-1},x_t) = u_t.
\end{equation}
If each $\pi_t$ only depends on the current state, i.e., $\pi_t(x_t) = u_t$, then $\pi$ is called a \emph{Markov policy}. Another form of $\pi_t$ is $\pi_t(x_t,y_t) = u_t$, where $y_t$ depends on prior states and actions (see Section \ref{statespaceaug} for an example).

Given $N \in \mathbb{N}$, often we consider a random cost $Z : \Omega \rightarrow \mathbb{R}$ of the form
\begin{align}\label{myZcum}
      Z & \coloneqq c_N(X_N) + \sum_{t = 0}^{N-1} c_t(X_t,U_t),
\end{align}
where $c_t : S \times C \rightarrow \mathbb{R}$ is a stage cost function for every $t \in \mathbb{T}$ and $c_N : S \rightarrow \mathbb{R}$ is a terminal cost function. One defines the stage and terminal cost functions to reflect the performance or safety criteria for the system of interest. For example, for a stormwater system, a stage or terminal cost function may quantify the volume of water that exceeds the capacity of the system's storage tanks.
For the treatment of blood cancer, a stage cost function may quantify the drug dosage, while a terminal cost function may assess the deviation between healthy blood cell counts and their desired ranges at the end of a treatment cycle.

Given $x \in S$ and $\pi \in \Pi$, the notation $E_{x}^\pi(Z)$ denotes the expectation of a random cost $Z$ that arises from the behaviour of the system \eqref{sysmodel} when the initial state is $x$ and the system uses the policy $\pi$. The form of $E_{x}^\pi(Z)$ depends on several items, including $x$, $\pi$, $q_t$ \eqref{qt}, $Z$, and the definition of the system's trajectory.\footnote{For details, we refer the interested reader to, for example, \cite[Prop. 7.28]{bertsekas2004stochastic} and \cite[Remark C.11]{hernandez2012discrete}.}

While different constructions are possible, the previous paragraphs describe some key building blocks that are useful for constructing a \emph{Markov decision process} (MDP) model.\footnote{We refer the reader to \cite[Sec. 2.2]{hernandez2012discrete}, \cite[Def. 8.1]{bertsekas2004stochastic}, and \cite[Def. 9.1]{bertsekas2004stochastic} for more technical presentations about the foundations of MDPs.} 
An MDP is a standard model in reinforcement learning, stochastic control, and operations research. The model can represent a variety of real systems and often leads to algorithms that can be implemented in a computer.

The term \emph{regularity conditions} refers to theoretical conditions that ensure that integrals, e.g., expectations, are well-defined and optimal policies exist. While regularity conditions vary according to the problem at hand, overall these conditions facilitate the analysis of fairly general systems (e.g., systems with uncountable state, action, and disturbance spaces, non-Gaussian disturbance distributions $p_t$, nonlinear dynamics functions $f_t$, and nonquadratic stage and terminal cost functions $c_t$). Examples of regularity conditions include the action space $C$ is compact, each $c_t$ is bounded below and continuous, and each $f_t$ is continuous. 

In Section \ref{emerging}, which focuses on risk-averse model-free methods, we consider an MDP with finitely many states and actions. Let us describe this model in more detail here. The disturbance distributions $p = p_t$, dynamics functions $f = f_t$, and stage cost functions $c=c_t$ do not vary with time $t$. It is convenient to enumerate the states as follows: $S = \{1,2,\dots,\ell\}$, where $\ell \in \mathbb{N}$ is given. We denote a state by $i$ or by $j$ to emphasize the countable nature of $S$. The quantity $q(\{j\}|i,u)$ is the probability that the realization of $X_{t+1}$ is 
$j \in S$, given that the realization of $(X_t,U_t)$ is $(i,u) \in S \times C$, i.e., 
%
\begin{subequations}\label{pinfinitecase}
\begin{equation}\label{myq}
    q(\{j\}|i,u) = p(\{ \hspace{.5mm} w \in D : f(i,u,w) \in \{j\} \hspace{.5mm} \}).
\end{equation}
The following notation is standard:
\begin{equation}\label{8b}
    \textbf{p}_{ij}(u) \coloneqq \mathbb{P}(X_{t+1} = j|X_t = i, U_t = u) \coloneqq q(\{j\}|i,u).
\end{equation}
Given a function $\pi : S \rightarrow C$, the sequence $(\pi,\pi,\dots)$ is a deterministic Markov policy.
The notation $P^\pi \in \mathbb{R}^{\ell \times \ell}$ denotes the matrix of conditional probabilities under the policy $(\pi,\pi,\dots)$. The $i$th row of $P^\pi$ is given by
\begin{equation}\label{matrixpi}
    P_i^\pi = \begin{bmatrix}\textbf{p}_{i1}(\pi(i)) & \textbf{p}_{i2}(\pi(i)) & \cdots &  \textbf{p}_{i\ell}(\pi(i))\end{bmatrix}, \;\;\;\;\;\ i \in S.
\end{equation}
\end{subequations}
It is common to use $P^\pi$ without defining $q$ \eqref{myq}. However, defining $q$ clarifies the relationship between $P^\pi$ and the model \eqref{sysmodel}.
Next, we will introduce different ways to summarize numerically the consequences that may occur due to a system’s behaviour.

\section{Methods for quantifying risk}\label{howtoquantifyrisk}
{Three concepts that are helpful for quantifying the risk of an autonomous system include
\begin{enumerate}
    \item probabilities of harmful events,
    \item temporal logic specifications, which can be deterministic or probabilistic, and
    \item risk functionals of random costs.
\end{enumerate}
First, we will introduce these concepts using an example. Then, in the rest of the section, we will emphasize the third concept.} Suppose that $Z$ is a random cost that arises from the behaviour of an autonomous system. 
%
%
For example, consider an urban stormwater system, which is a network of artificial and natural infrastructure (e.g., pipes, tanks, ponds, and streams) near a city that collects, distributes, or treats stormwater. Suppose that automated valves are located throughout the network to manage the flow of water, and the network can release excess flows into a combined sewer.\footnote{Combined sewers can be found in older cities, including San Francisco and Toronto. In prior work, we developed a risk-averse safety analysis method, which we applied to the problem of quantifying the risk of combined sewer overflows \cite{mpctacsubmission, chapman2021risk}.} An example of $Z$ is the random total volume of water to be discharged into the combined sewer next month. We prefer smaller realizations of $Z$ because a combined sewer can discharge a mixture of stormwater and untreated wastewater into the environment. For this example, we illustrate the concepts of a harmful event, a temporal logic specification, and a risk functional of $Z$ in Table \ref{examplewater}. 

\begin{table}[h!]
\centering
\begin{center}
\begin{tabular}{ |p{70pt}|p{175pt}|p{190pt}|} 
 \hline
{\bf {Concept}} & {\bf {Example of the concept}} & {\bf {Example problem of interest using the concept}}\\
\hline 
Harmful event & The event that the random water volume $Z$ exceeds a given threshold $z > 0$ is denoted by $\{Z \geq z\}$. &  Develop an algorithm to operate the automated valves in a manner that minimizes the probability of $\{Z \geq z\}$.\\
\hline
Temporal logic specification & Until the second storm of the month, the probability that the combined sewer discharges more than $z$ m$^3$ of water is less than 0.05. 
&  Find a policy for the automated valves that minimizes their total power consumption subject to the specification.\\ 
 \hline
 Risk functional $\rho$ & Let $\rho(Z)$ be the average value of $Z$ in the $1\%$ of the worst cases based on estimates from historical precipitation data. & Design a policy for the valves that minimizes $\rho(Z)$. Determine the minimum capacity of the system so that $\rho(Z) \leq z$ holds when the valves are open, where $z$ is a given threshold.\\
 \hline
\end{tabular} 
\end{center}
\caption{In the context of an urban water system with automated valves, we illustrate the concepts of a harmful event, a temporal logic specification, and a risk functional of a random cost $Z$. Here, $Z$ is the random total volume of water to be released by the system into a combined sewer next month. We prefer smaller realizations of $Z$ because such realizations correspond to smaller volumes of untreated wastewater that could pollute the environment.}
\label{examplewater}
\end{table}

{Temporal logic is a mathematical language for describing relationships between different events in time and was originally proposed for verifying computer programs in the 1970s \cite{pnueli1977temporal}. The modern literature on temporal logic and its application to verifying the behaviour of an autonomous system is extensive, deserving a full survey of its own. 
While a detailed discussion is out of scope, we illustrate temporal logic specifications in Table \ref{stl}, and we refer the reader to the works, e.g., \cite{coogan2017formal, kwiatkowska2007stochastic, forejt2011automated}, for comprehensive presentations.}
\begin{table}[h!]
\centering
\begin{center}
\begin{tabular}{ |p{460pt}|} 
\hline
{\textbf{Examples of deterministic temporal logic specifications}:} \\

$\bullet$ The system remains in a region $K$ always and reaches a target $T \subseteq K$ eventually.\\
 $\bullet$ The system visits a given region of the state space infinitely often and remains in a region $K$ until a specific event occurs. \\
$\bullet$ The system visits a region $K_1$ after visiting another region $K_2$ always.\\
 \hline
 {\textbf{Examples of probabilistic temporal logic specifications}:} 
 \\ $\bullet$ The system remains in a region $K$ with probability at least 0.9 and reaches a target $T \subseteq K$ eventually with probability at least 0.8.\\
$\bullet$ The probability that the system causes a particular event to occur within the next hour is less than 0.01. \\
$\bullet$ If the system visits a region $K_1$, then it enters another region $K_2$ in the next minute with probability at least 0.95.\\
\hline
\end{tabular} 
\end{center}
\caption{Examples of deterministic and probabilistic temporal logic specifications for autonomous systems.}
\label{stl}
\end{table}

\begin{figure}[!h]
\centerline{\includegraphics[width=0.6\textwidth]{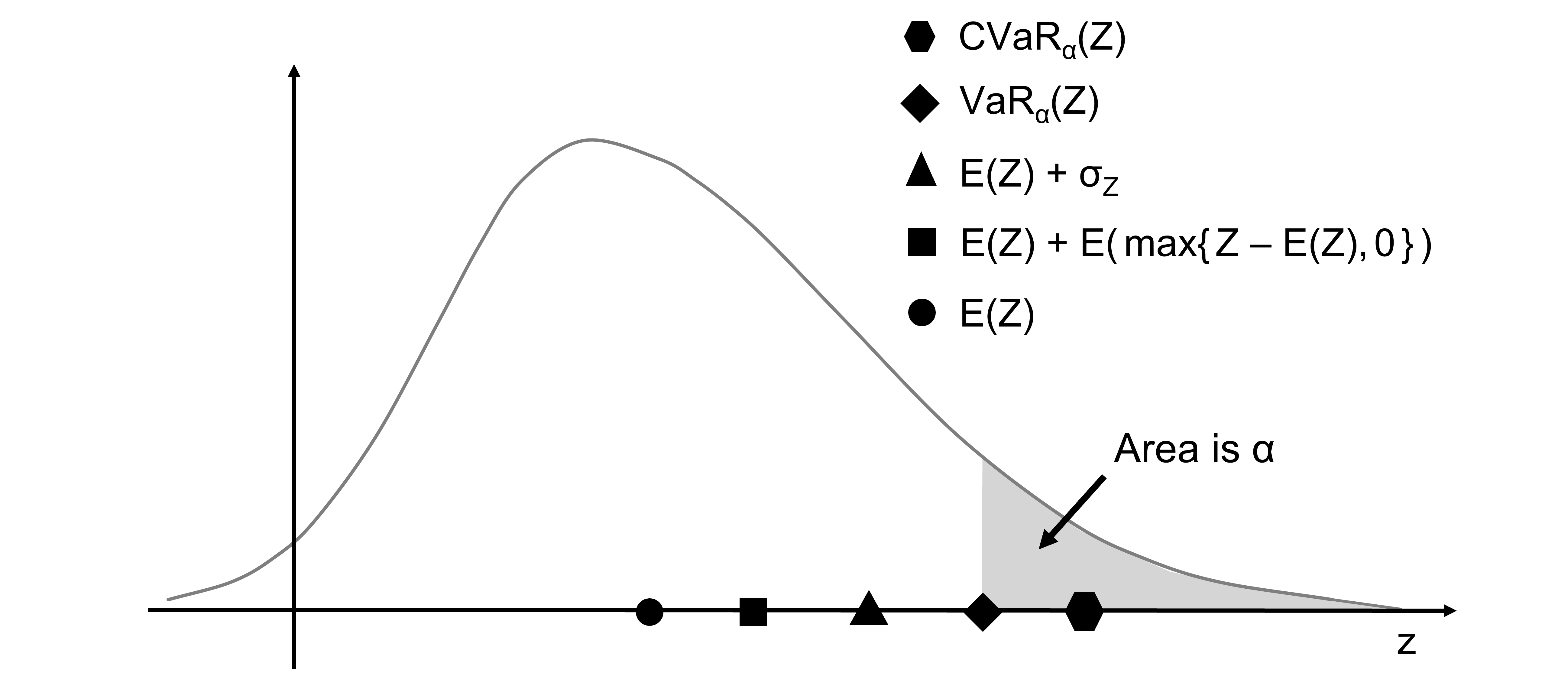}}
\caption{A probability density of a random cost $Z$; recall from the Introduction that a random cost is a random variable for which smaller realizations are preferred. We depict different characteristics of the distribution of $Z$: $E(Z)$ is the expectation of $Z$; $E(\max\{Z-E(Z), 0\})$ is the expectation of $\max\{Z-E(Z), 0\}$; {$\sigma_Z$ is the standard deviation of $Z$}; $\text{VaR}_\alpha(Z)$ is the left-side $(1-\alpha)$-quantile of $Z$, which is called the value-at-risk of $Z$ at level $\alpha$; and $\text{CVaR}_\alpha(Z)$ is the conditional value-at-risk of $Z$ at level $\alpha$. Informally, $\text{CVaR}_\alpha(Z)$ represents the average value of $Z$ above $\text{VaR}_\alpha(Z)$; we will provide details in Section \ref{mycvarsection}. In this figure, we depict a setting in which $Z$ admits a density and $\alpha \in (0,1)$ is small enough so that $E(Z) + \sigma_Z < \text{VaR}_\alpha(Z)$.}
\label{differetcharacteristicsofdistribution}
\end{figure}

{The rest of this section is about the theory of risk functionals. A \emph{risk functional} is a map $\rho$ from a space of random variables to $\mathbb{R}\cup\{+\infty\}$.\footnote{It is convenient to not permit risk functionals to return $-\infty$ so that a sum of risk functionals is well-defined.} That is, the domain of $\rho$ is a space of random variables, and the codomain of $\rho$ is $\mathbb{R}\cup\{+\infty\}$. More specifically, the domain is the space of random variables 
with finite $\text{p}$\textsuperscript{th} moment for a fixed $\text{p} \in [1,+\infty)$, and we 
assume that there is a random variable $Z \in L^\text{p}$ so that $\rho(Z)$ is finite, 
following convention \cite{shapiro2009lectures}. We prefer smaller realizations of a random variable, meaning that the realizations represent costs. 
The term risk measure is a synonym for risk functional.\footnote{The term risk functional emphasizes that the domain of $\rho$ is a space of functions. The term risk measure emphasizes that the purpose of $\rho$ is to assess risk. The latter term should not be confused with a probability measure, whose domain is a sigma algebra.} We have already seen some examples of risk functionals, including the mean, a sum of the mean and variance, an expected utility, a quantile, and an average cost above a quantile. The terms ``quantile'' and ``average cost above a quantile'' are somewhat informal. We will present these terms more formally by defining the value-at-risk (VaR) and the conditional value-at-risk (CVaR), respectively. Figure \ref{differetcharacteristicsofdistribution} illustrates VaR, CVaR, and some other risk functionals for a random variable that has a density.}
%

The modern axiomatic theory of risk functionals has been developed by the operations research and financial engineering communities. While the theory has been motivated by the challenge of managing financial portfolios, it applies to autonomous systems operating under uncertainty more generally. In Section \ref{examplesriskfunctionals} and Section \ref{extremalriskfunctionals}, we will present examples of risk functionals. Then, in Section \ref{advdis}, we will discuss their advantages and disadvantages. 

\subsection{Examples of risk functionals}\label{examplesriskfunctionals}
In 1994, Heger introduced several risk functionals to the machine learning community, including expected utility, mean-variance, and a functional that resembles value-at-risk \cite{heger1994consideration}. These functionals are popular today as well. More broadly, common classes of risk functionals include utility-based functionals, functionals that quantify deviations from the mean, quantile-based functionals, and nested risk functionals. Most of the risk functionals that we present in Section \ref{examplesriskfunctionals} are also presented by \cite[Sec. 6.3.2]{shapiro2009lectures}, from which we take inspiration. However, our presentation is less technical and provides connections to the autonomous systems literature. {Throughout the section, we work with an arbitrary probability space $(\Omega,\mathcal{F},\mu)$ and a random variable $Z : \Omega \rightarrow \mathbb{R}$. Expectation $E(\cdot)$ is defined with respect to the probability measure $\mu$. The cumulative distribution function (CDF) of $Z$ is defined by $F_Z(z) \coloneqq \mu(\{Z \leq z \})$ for every $z \in \mathbb{R}$.} 
%
%
\subsubsection{Expected utility}\label{expectedutility}
The notion of a utility function is based on the idea that two people can view the same outcome, event, or reward differently \cite[Ch. 1]{eeckhoudt2011economic}. This notion was first studied in the literature by D. Bernoulli in 1738, whose paper was translated from Latin to English in 1954 \cite{Bernoulli}. von Neumann and Morgenstern analyzed the use of expected utility functionals to model subjective preferences from a decision-theoretic perspective in 1944 \cite{von1948theory}. 

{The \emph{expected utility} of a nonnegative random cost $Z$ takes the form $E(\phi(Z))$, where $\phi : \mathbb{R}_{+} \rightarrow \mathbb{R}$ is a nondecreasing continuous function. $\phi$ is called a \emph{utility function} and is chosen to represent a decision-maker's subjective preferences. 
If $\phi$ is (strictly) convex, then $\phi(z)$ represents how a risk-averse decision-maker perceives a realization $z \in \mathbb{R}_+$ of the random cost $Z$ \cite{eeckhoudt2011economic, bauerlerieder}.
%
%
%
%
{Convexity is associated with risk aversion because if $\phi$ is convex, then $\phi(E(Z)) \leq E(\phi(Z))$ by Jensen's Inequality, and the nonrandom outcome $\phi(E(Z))$ is preferred to the random outcome in expectation $E(\phi(Z))$ \cite[p. 7]{eeckhoudt2011economic}.}\footnote{{Suppose that a random cost $\tilde{Z}$ equals $E(Z)$ with probability one. Then, the expected utility of $\tilde{Z}$ is given by $E(\phi(\tilde{Z})) = 1 \cdot \phi(E(Z)) = \phi(E(Z))$. If $\phi$ is convex, then we have $\phi(E(Z)) \leq E(\phi(Z))$, and therefore, we conclude that $E(\phi(\tilde{Z})) \leq E(\phi(Z))$. From the perspective of a risk-averse decision-maker, $\tilde{Z}$ has a lower expected utility compared to $Z$, and thus, it is preferable.}}}
It is common to assess a related quantity 
\begin{equation}\label{rhophi}
     \rho_{\phi}(Z) \coloneqq \phi^{-1}\big(E(\phi(Z))\big),
\end{equation}
provided that $\phi$ is strictly increasing and continuous and its inverse $\phi^{-1}$ exists \cite[p. 107]{bauerlerieder}. The inverse operation ensures that the units of $\rho_{\phi}(Z)$ are equivalent to the units of the realizations of $Z$. This may help interpret $\rho_{\phi}(Z)$. For simplicity, we call $\rho_{\phi}$ an expected utility functional as well.
\subsubsection{Exponential utility}
The exponential utility functional of a nonnegative random cost $Z$ at level $\theta \in \mathbb{R}$ with $\theta \neq 0$,
\begin{equation}\label{eu}
    \rho_{\text{e},\theta}(Z) \coloneqq \textstyle \frac{-2}{\theta} \log E(\exp(\frac{-\theta}{2}Z)),
\end{equation}
is a special case of $\rho_{\phi}(Z)$ \eqref{rhophi} with $\phi(z) = \exp(\frac{-\theta}{2}z)$. A synonym for exponential utility is the entropic risk functional. Typically, $\rho_{\text{e},\theta}(Z)$ is interpreted as an approximation for a linear combination of the mean $E(Z)$ and variance {$\sigma_Z^2$} of $Z$. Under appropriate conditions, it can be shown that
\begin{equation}\label{1}
    \lim_{\theta \rightarrow 0^+} \rho_{\text{e},\theta}(Z) = \lim_{\theta \rightarrow 0^-} \rho_{\text{e},\theta}(Z) = E(Z),
\end{equation}
and following \cite[p. 765]{whittle1981}, if the magnitude of $\theta \cdot \sigma_Z^2$ is small enough, then
\begin{equation}\label{myapprox}
   \rho_{\text{e},\theta}(Z) \approx E(Z) - \textstyle \frac{\theta}{4} \cdot {\sigma_Z^2}.
\end{equation}
We provide sufficient conditions to derive \eqref{1} in a footnote.\footnote{Suppose that $Z \in L^2$, $Z$ is nonnegative, and there exist $a \in \mathbb{R}$ and $b \in \mathbb{R}$ such that $a < 0 < b$ and $\exp(\frac{-\theta}{2}Z) \in L^2$ for every $\theta \in [a,b]$. One can use \cite[Th. 2.27, p. 56]{folland1999real} and H\"{o}lder's Inequality to show that $\frac{\mathrm{d}}{\mathrm{d}\theta}E(\exp(\frac{-\theta}{2}Z)) = \frac{-1}{2} E(Z \exp(\frac{-\theta}{2}Z))$, $\underset{\theta \rightarrow 0}{\lim} E(Z \exp(\frac{-\theta}{2}Z)) = E(Z)$, and $\underset{\theta \rightarrow 0}{\lim} E(\exp(\frac{-\theta}{2}Z)) = 1$. Then, one completes the proof of \eqref{1} using L'H\^{o}pital's Rule.\label{footnote6}}
{In the case of $\theta < 0$, the variance is added to the mean in the approximation \eqref{myapprox}, representing a preference for small variance; this is an example of a risk-averse perspective. However, in the case of $\theta > 0$, the variance is subtracted from the mean in the approximation \eqref{myapprox}, representing a preference for large variance.}
We will discuss the optimization of exponential utility performance criteria for MDPs in Section \ref{exponentialutilitymdps}.
\subsubsection{Mean-variance, mean-deviation, and mean-upper-semideviation} Here, we describe different ways to quantify the spread of a distribution relative to the mean. Let $\text{p} \in [1,+\infty)$, $Z \in L^\text{p}$, and $\lambda \geq 0$ be given. 
The \emph{mean-variance} of $Z \in L^2$ is a linear combination of the mean and variance of $Z$, i.e.,
\begin{equation}\label{mv}
    \rho_{\text{MV},\lambda}(Z) \coloneqq E(Z) + \lambda \cdot {\sigma_Z^2}.
\end{equation}
It is typical to choose $Z \in L^2$ so that the variance 
is finite.
Assessing mean-variance trade-offs has an extensive history in the financial literature. The famous works by H. Markowitz from the 1950s focus on portfolio selection under competing mean-variance objectives and justify using variance as a measurement for financial risk \cite{Markowitz1952, harry1959portfolio}. More recently, this foundation has motivated the development of methods for autonomous systems that are designed to
\begin{enumerate}
    \item minimize the variance such that the mean must equal a particular number, which is an inexpensive computation if the system has linear dynamics and quadratic costs \cite{won2005cost}, and
    \item minimize a linear combination of the mean and variance, which is typically an expensive optimization problem \cite[Ex. 1]{milleryang}.
\end{enumerate}

The \emph{mean-deviation} of $Z$ is a linear combination of the mean of $Z$ and the $L^\text{p}$-norm of $Z - E(Z)$, i.e.,
\begin{equation}\label{md}
    \rho_{\text{MD},\lambda}(Z) \coloneqq E(Z) + \lambda \|Z - E(Z)\|_\text{p}.
\end{equation}
For example, if $\text{p} = 2$, then $\rho_{\text{MD},\lambda}(Z)$ is a linear combination of the mean and standard deviation of $Z$:
\begin{equation}
    \rho_{\text{MD},\lambda}(Z) = E(Z) + \lambda \|Z - E(Z)\|_2 =E(Z) + \lambda \cdot {\sigma_Z}. 
\end{equation}
The \emph{mean-upper-semideviation} of $Z$ is a weighted sum of $E(Z)$ and the $L^\text{p}$-norm of $\max\{Z - E(Z),0\}$, i.e.,
\begin{equation}\label{meanupper}
    \rho_{\text{MU},\lambda}(Z) \coloneqq E(Z) + \lambda \|\max\{Z - E(Z),0\}\|_\text{p}.
\end{equation}
The mean-upper-semideviation of $Z$ \eqref{meanupper} assesses positive deviations of $Z$ relative to $E(Z)$. In contrast, the mean-variance \eqref{mv} and mean-deviation \eqref{md} do not differentiate between positive and negative deviations of $Z$ relative to $E(Z)$.

%
\subsubsection{Value-at-risk} 
\begin{figure}[!h]
\centerline{\includegraphics[width=0.85\textwidth]{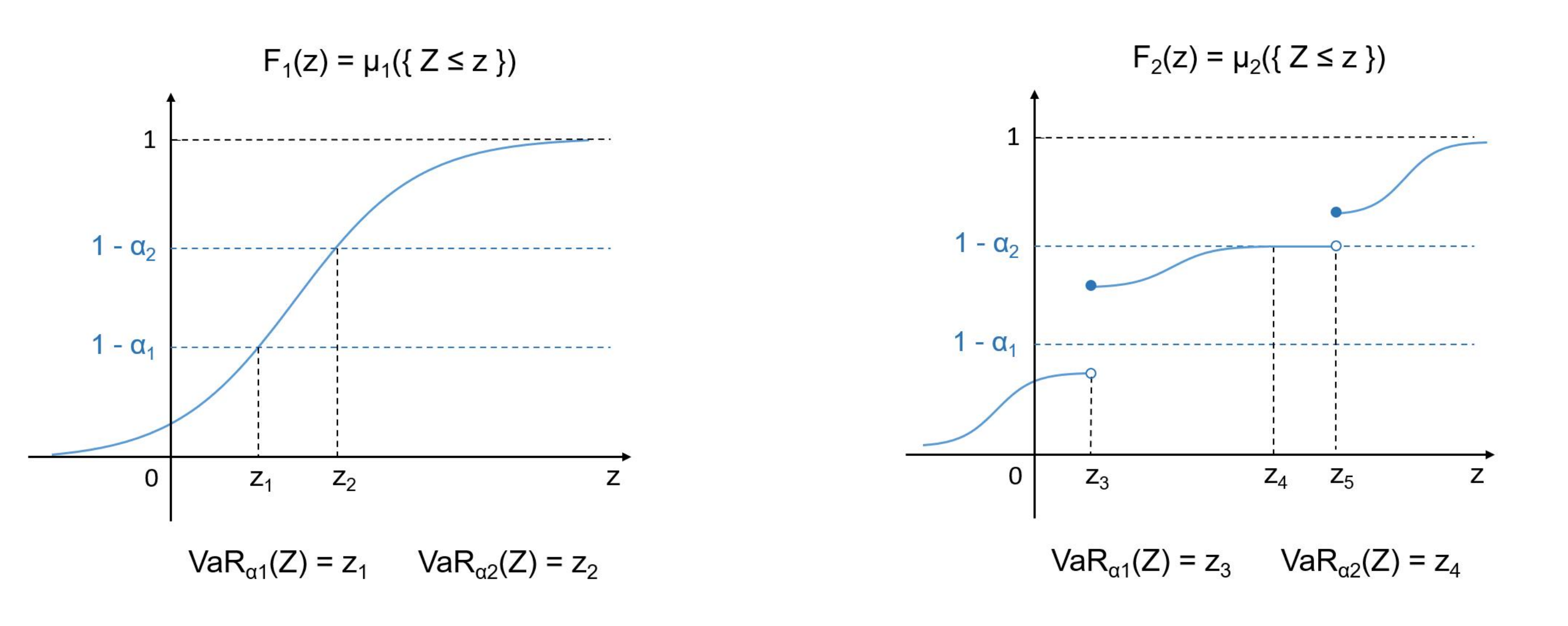}} 
\caption{An illustration of the value-at-risk (VaR) of a random cost $Z \in L^1(\Omega,\mathcal{F},\mu_i)$. To explain the VaR, we consider two distributions $F_i(z) \coloneqq \mu_i(\{Z \leq z\})$ for $i = 1,2$. $F_1$ is continuous and strictly increasing (left panel), while $F_2$ is right-continuous and nondecreasing (right panel). $\mu_i$ is a probability measure, $F_i$ is the CDF that corresponds to $\mu_i$, $0 < \alpha_2 \leq \alpha_1 < 1$, and $z_j$ is a realization of $Z$. On the left, it holds that $F_1(z_1)=1-\alpha_1$ and $F_1(z_2)=1-\alpha_2$. Thus, when the VaR is defined using $\mu_1$, it holds that $\text{VaR}_{\alpha_1}(Z)=z_1$ and $\text{VaR}_{\alpha_2}(Z)=z_2$. On the right, there is no $z$ such that $F_2(z)=1-\alpha_1$, but for any $z \geq z_3$, it holds that $F_2(z) \geq 1-\alpha_1$. Consequently, when the VaR is defined using $\mu_2$, we have that $\text{VaR}_{\alpha_1}(Z)=z_3$. Also, on the right, it holds that $F_2(z)=1-\alpha_2$ for every $z \in [z_4,z_5)$. Since $z_4$ is the smallest value in $[z_4,z_5)$, it follows that $\text{VaR}_{\alpha_2}(Z)=z_4$ when the VaR is defined using $\mu_2$.}
\label{varfig}
\end{figure}

{We have introduced the value-at-risk in the context of a random variable that has a density in Figure \ref{differetcharacteristicsofdistribution}. We use the notation $\text{VaR}_\alpha(Z)$ to denote 
the value-at-risk of a random variable $Z$ at level $\alpha \in (0,1)$; $F_Z$ denotes the cumulative distribution function of $Z$. $\text{VaR}_\alpha(Z)$ is the left-side $(1-\alpha)$-quantile of the distribution of $Z$. Equivalently, $\text{VaR}_\alpha(Z)$ is the generalized inverse of $F_Z$ at level $1-\alpha$, which we depict in Figure \ref{varfig}.}  
%

{Now that we have illustrated the VaR, we are ready to present its mathematical definition. The value-at-risk of a random variable $Z$ at level $\alpha$ is defined by}
\begin{equation}\label{myvar}
    \text{VaR}_\alpha(Z) \coloneqq \inf\{z \in \mathbb{R} : F_Z(z) \geq 1 - \alpha \}, \;\;\;\;\;\alpha \in (0,1).
\end{equation}
Recall that $F_Z$ is right-continuous and nondecreasing \cite[p. 209]{ash1972probability}. 
For a given $\alpha \in (0,1)$, there may be different points $z \neq z'$ such that $F_Z(z) = F_Z(z') = 1 - \alpha$, or there may be no such point (Figure \ref{varfig}). 
$\text{VaR}_\alpha(Z)$ \eqref{myvar} represents the smallest value of the set $\{z \in \mathbb{R} : F_Z(z) \geq 1 - \alpha \}$.
%
\subsubsection{Conditional value-at-risk}\label{mycvarsection} 
The conditional value-at-risk of $Z$ represents the expectation of $Z$ in a given fraction of the worst cases (Figure \ref{differetcharacteristicsofdistribution}). CVaR was developed in its modern form in the early 2000s, and the key works underlying this development include \cite{Rockafellar2000, rockafellar2002conditional, acerbitasche}. Synonyms for CVaR are average value-at-risk, expected shortfall, and expected tail loss \cite{shapiro2012}. CVaR has various representations, and these representations offer different insights about how this functional characterizes a distribution. 

First, we will introduce the representation that explains the name conditional value-at-risk. Suppose that $Z \in L^1$, $\alpha \in (0,1)$, and $F_Z$ is continuous at the point $z=\text{VaR}_\alpha(Z)$. Under these conditions, $\text{CVaR}_\alpha(Z)$ is equivalent to the expectation of $Z$ conditioned on the event that $Z$ exceeds $\text{VaR}_\alpha(Z)$ \cite[Th. 6.2]{shapiro2009lectures}. 
This representation is depicted in Figure \ref{differetcharacteristicsofdistribution} in the special case when $Z$ has a density. 

The next representation to be described explains the name average value-at-risk. If $Z \in L^1$ and $\alpha \in (0,1)$, then $\text{CVaR}_\alpha(Z)$ equals an average of the value-at-risk as follows:
\begin{equation}\label{12}
    \text{CVaR}_{\alpha}(Z) = 
 \frac{1}{\alpha} \int_{1-\alpha}^1 \text{VaR}_{1-s}(Z) \; \mathrm{d}s.
\end{equation}
In \eqref{12}, the integral is taken with respect to the level of the VaR.
A proof is provided by \cite[Th. 6.2]{shapiro2009lectures}. 

The last representation often serves as the definition of CVaR. This representation expresses CVaR in terms of a convex optimization problem. 
If $Z \in L^1$ and $\alpha \in (0,1]$, then $\text{CVaR}_{\alpha}(Z)$ is defined by
\begin{equation}\label{cvardefdef}
    \text{CVaR}_{\alpha}(Z) \coloneqq  \underset{s \in \mathbb{R}}{\inf} \Big( s + \textstyle\frac{1}{\alpha} E(\max\{Z-s,0\}) \Big).
\end{equation} 
In particular, one can use \eqref{cvardefdef} to prove that $\text{CVaR}_{1}(Z)$ equals the expectation of $Z$. For any $\alpha \in (0,1)$, a minimizer of the right side of \eqref{cvardefdef} is $\text{VaR}_{\alpha}(Z)$,\footnote{For instance, see \cite[p. 258]{shapiro2009lectures} or \cite[p. 721]{shapiro2012}.} and therefore,
\begin{equation}\label{encoderiskcvar}
    \text{CVaR}_{\alpha}(Z) = \text{VaR}_{\alpha}(Z) + \textstyle\frac{1}{\alpha} E(\max\{Z-\text{VaR}_{\alpha}(Z),0\}).
\end{equation}
Equation \eqref{encoderiskcvar} means that the CVaR at level $\alpha \in (0,1)$ characterizes the distribution of $Z$ in terms of a weighted sum of $\text{VaR}_{\alpha}(Z)$ and the average amount that $Z$ exceeds $\text{VaR}_{\alpha}(Z)$. We will describe a method to solve a risk-averse optimal control problem, where the objective is defined using CVaR, in Section \ref{statespaceaug}. 

\subsubsection{Nested (i.e., recursive, compositional) risk functionals} \label{nested} 
In this section, we will present a nested risk functional on a short time horizon for simplicity, and then, we will present a more general setting.
Let $Z_j$ be a random cost corresponding to time $j \in \{0,1,2\}$, and consider the random cumulative cost $Z = Z_0 + Z_1 + Z_2$. A nested risk functional $\rho$ of $Z$ takes the form
\begin{equation}\label{mynesteq1}
    \rho(Z) = \nu_0(Z_0+ \nu_1(Z_1 + \nu_2(Z_2))),
\end{equation}
where each $\nu_t : \mathcal{Z}_{t+1} \rightarrow \mathcal{Z}_t$ is a map between spaces of random variables. The random variables in $\mathcal{Z}_{j}$ are permitted to depend on a process (e.g., a system's random state trajectory) from time 0 to time $j$. In this formulation, $\nu_t$ is called a \emph{one-step conditional} risk functional \cite{ruszczynski2010risk, ruszczynski2014erratum}. Another formulation is to define each $\nu_t$ to be a risk functional, i.e., a map from a space of random variables to $\mathbb{R} \cup \{+\infty\}$ \cite{bauerle2021markov}. 

In \eqref{mynesteq1}, the time horizon has length $N = 2$. More generally, we may consider a time horizon of length $N \in \mathbb{N}$. In this case, a nested risk functional $\rho$ of a random cost $Z = Z_0 + Z_1 + \dots + Z_N$ takes the form
\begin{equation}\label{gennested}
    \rho(Z) = \nu_0(Z_0+ \nu_1(Z_1 + \dots + \nu_{N-1}(Z_{N-1} + \nu_N(Z_N)) \cdots)).
\end{equation}
The problem of minimizing a random cost that is assessed using a nested risk functional for an autonomous system has been studied by several authors over the past decade. Example works between 2010 and 2022 include \cite{ruszczynski2010risk, ruszczynski2014erratum, bauerle2021markov, shen2013risk, singh2018framework, kose2021risk}. We will present a model-based method from \cite{bauerle2021markov} in Section \ref{methodnested} and a model-free method from \cite{kose2021risk} in Section \ref{emerging}.

\subsection{Extremal risk functionals}\label{extremalriskfunctionals}
Some risk functionals from the previous sections can be classified as \emph{extremal risk functionals} \cite{milleryang}. A risk functional in this class takes the form
\begin{equation}\label{ext}
    \rho_{\text{m}}(Z) = \inf_{s \in \mathbb{R}^\text{q}} E\big(g(Z,s)\big),
\end{equation}
where $g : \mathbb{R} \times \mathbb{R}^\text{q} \rightarrow \mathbb{R}$ is a convex function and $\text{q} \in \mathbb{N}$. Let us discuss two examples from \cite{milleryang}. CVaR \eqref{cvardefdef} is an extremal risk functional with 
\begin{equation}
g(z,s) = s + {\textstyle\frac{1}{\alpha}}\max\{z-s,0\},
\end{equation}
$Z \in L^1$, and $\alpha \in (0,1]$ because
\begin{equation}
    \text{CVaR}_\alpha(Z) = \inf_{s \in \mathbb{R}} E\big(s + {\textstyle\frac{1}{\alpha}} \max\{Z-s,0\}\big).
\end{equation}
A second example is the mean-variance functional \eqref{mv} with 
\begin{equation}
    g(z,s) = z + \lambda (z - s)^2,
\end{equation}
$Z \in L^2$, and $\lambda \geq 0$ because
\begin{equation}
    E(Z) + \lambda \cdot \sigma_Z^2 = E(Z) +  \lambda \cdot \inf_{s \in \mathbb{R}} E\big( (Z - s)^2 \big) = \inf_{s \in \mathbb{R}} E\big( Z + \lambda  (Z - s)^2 \big). 
\end{equation}
The first equality holds because
\begin{equation}
    \inf_{s \in \mathbb{R}} E\big( (Z - s)^2 \big) = \inf_{s \in \mathbb{R}} \Big( E(Z^2) - 2 s \cdot E(Z) + s^2 \Big),
\end{equation}
and therefore, the objective is a convex quadratic function of $s$ and has a unique minimizer $s^* = E(Z)$. Other examples of extremal risk functionals include a linear combination of the mean and CVaR and a linear combination of the variance and CVaR \cite{milleryang}.

\subsection{Advantages and disadvantages}\label{advdis}
Having presented different risk functionals, next we will discuss their respective advantages and disadvantages in terms of decision-theoretic axioms, computational tractability, and explainability.\footnote{{The term \emph{computational tractability} refers to the amount of resources (e.g., time, energy, and memory) that is required to implement an algorithm in a computer. In this survey, we typically discuss this notion in a qualitative sense. For example, suppose that at time $t \in \mathbb{N}$, Algorithm $a$ requires knowledge of the system's trajectory from time zero to time $t$. In contrast, Algorithm $b$ requires knowledge of the system's state at time $t$, but it does not need a record of the trajectory prior to time $t$. Then, Algorithm $b$ is more tractable than Algorithm $a$.}}

\subsubsection{Axiomatic considerations} Some risk functionals enjoy a strong axiomatic foundation, including 
\begin{itemize}
    \item mean-upper-semideviation $\rho_{\text{MU},\lambda}$ \eqref{meanupper} on the domain $L^\text{p}$ with $\text{p} \in [1,+\infty)$ and $\lambda \in [0,1]$ \cite[Ex. 6.20]{shapiro2009lectures} and
    \item CVaR$_{\alpha}$ \eqref{cvardefdef} on the domain $L^1$ with $\alpha \in (0,1]$.
\end{itemize}
Specifically, these two functionals satisfy four properties that define the class of \emph{coherent} risk functionals: \emph{monotonicity}, \emph{subadditivity}, \emph{translation equivariance}, and \emph{positive homogeneity}.\footnote{Here, we present these properties formally. Let $Z_i \in L^\text{p}(\Omega,\mathcal{F},\mu)$, $\text{p} \in [1,+\infty)$, $a \in \mathbb{R}$, $\beta \in \mathbb{R}_+$, and $\rho : L^\text{p} \rightarrow \mathbb{R} \cup\{+\infty\}$ be given. The abbreviation $\mu$-a.e. specifies a property that holds almost everywhere with respect to $\mu$. The risk functional $\rho$ having the domain $L^\text{p}$ means that $\rho$ evaluates random variables whose $\text{p}$\textsuperscript{th} moments are finite and if $Z_1$ equals $Z_2$ $\mu$-a.e., then $\rho(Z_1) = \rho(Z_2)$. $\rho$ is \emph{monotonic} if and only if $Z_1 \leq Z_2$ $\mu$-a.e. implies that $\rho(Z_1) \leq \rho(Z_2)$. $\rho$ is \emph{subadditive} if and only if $\rho(Z_1 + Z_2) \leq \rho(Z_1) + \rho(Z_2)$. $\rho$ is \emph{translation-equivariant} if and only if $\rho(Z_1 + a) = \rho(Z_1) + a$. $\rho$ is \emph{positively homogeneous} if and only if $\rho(\beta Z_1) = \beta \rho(Z_1)$.} The class of coherent risk functionals was originally proposed by Artzner et al. \cite{Artzner}, which justified the importance of the four properties in financial applications. Also, these properties can be justified more broadly. Monotonicity means that a random cost that is larger than another almost everywhere incurs a larger risk. This is a logical property because larger costs indicate more harm. Subadditivity $\rho(Z_1 + Z_2) \leq \rho(Z_1) + \rho(Z_2)$ means that ``a merger does
not create extra risk'' in financial engineering \cite{Artzner}. More generally, a ``merger'' can be viewed as a system, network, or sequential process of nonadversarial interacting components, suggesting applications in robotics \cite{majumdar2020should}, healthcare, renewable energy, and natural resources management. The last two properties ensure that the risk of a translated or scaled random cost is related to the risk of the untransformed random cost in a natural way (based on the specific translation or scaling). Unfortunately, value-at-risk, exponential utility, and mean-variance are not coherent. Value-at-risk \eqref{myvar} is not subadditive \cite{Artzner, kisi}. Exponential utility \eqref{eu} is not positively homogeneous.\footnote{For example, consider $\rho_{\text{e},\theta}(Z)$ \eqref{eu} with $\theta = -2$, where $Z$ equals 1 with probability 0.2 and $Z$ equals 2 with probability 0.8. If $\beta = 2$, then $\rho_{\text{e},\theta}(\beta Z) \approx 3.81$ but $\beta\rho_{\text{e},\theta}( Z) \approx 3.73$.} Mean-variance \eqref{mv} with $\lambda > 0$ is not positively homogeneous and not monotonic.\footnote{Consider $\rho_{\text{MV},\lambda}(Z_1)$ and $\rho_{\text{MV},\lambda}(Z_2)$ \eqref{mv} with $\lambda = 4$, where $Z_1$ is uniformly distributed on $(-1,1)$ and $Z_2$ equals 1.2 with probability 1. If $\beta = 2$, then $\rho_{\text{MV},\lambda}(\beta Z_1) = \frac{16}{3}$, but $\beta\rho_{\text{MV},\lambda}(Z_1) = \frac{8}{3}$. While $Z_1 \leq Z_2$ everywhere, it holds that $\rho_{\text{MV},\lambda}(Z_1) = \frac{4}{3} > \rho_{\text{MV},\lambda}(Z_2) = 1.2$.} 

\subsubsection{Computational tractability}
If $Z$ \eqref{myZcum} is a random cumulative cost incurred by an MDP, then the expectation of $Z$ and exponential utility of $Z$ can be optimized by formulating dynamic programs on the state space $S$ (to be described in Section \ref{exponentialutilitymdps}). A nested risk functional of $Z$ can be optimized using a dynamic program on $S$ as well (to be presented in Section \ref{methodnested}). However, optimizing performance criteria defined by other risk functionals, including CVaR, mean-variance, mean-CVaR, and expected (nonexponential) utility, require more expensive computations in general \cite{chapman2021risk,bauerlerieder,milleryang, pflug2015,bauerle2011markov,haskell,bauerle2020minimizing, kevinpaper}. 
One may need to solve a series of subproblems, where each subproblem depends on the history of the MDP from time zero. Hence, great computational resources may be required, e.g., a long runtime, large energy and memory requirements, and many cores. In contrast, each subproblem corresponding to the optimization of the expectation or exponential utility of $Z$ \eqref{myZcum} depends on the history only through the current state, which is more tractable. 

In prior empirical work, we investigated the two problems of optimizing the exponential utility of $Z$ \eqref{myZcum} and the CVaR of $Z$ for two systems in simulation \cite{kevinpaper}. We considered the following systems: 
\begin{enumerate}
    \item a thermostatic regulator with $S \subset \mathbb{R}$ and a time horizon of length $N = 12$ and
    \item a stormwater system with $S \subset \mathbb{R}^2$ and a time horizon of length $N = 48$.
\end{enumerate}
While our code was not designed specifically for efficiency, a qualitative comparison of resources highlights the (unsurprising) disadvantage of MDP problems that require history-dependent algorithms. The computations for the thermostatic regulator ($S \subset \mathbb{R}$) required 2 minutes for exponential utility and 25 minutes for CVaR on 4 cores \cite[Table IV]{kevinpaper}. The computations for the stormwater system ($S \subset \mathbb{R}^2$) required 10 minutes for exponential utility and 136 hours for CVaR on 30 cores \cite[Table IV]{kevinpaper}. 

\subsubsection{Explainability} 
{The conditional value-at-risk, value-at-risk, and mean-variance functionals enjoy intuitive interpretations, whereas the interpretations of other risk functionals may be less straightforward.} 
In particular, if $Z$ is a random variable with finite first moment and its cumulative distribution function is continuous, then the CVaR of $Z$ at level $\alpha \in (0,1)$ is the expectation of $Z$ in the $\alpha \cdot 100$\% worst cases. The parameter $\alpha$ in the definition of CVaR \eqref{cvardefdef} has a quantitative interpretation as a fraction of the most harmful realizations of $Z$, which we have illustrated when $Z$ admits a density in Figure \ref{differetcharacteristicsofdistribution}. In contrast, the parameter $\theta$ in the definition of exponential utility \eqref{eu} has a qualitative interpretation as a weight on the variance, provided that the mean-variance approximation \eqref{myapprox} is valid. 
More generally, the class of expected utility functionals has a qualitative interpretation of representing the subjective preferences of decision-makers (Section \ref{expectedutility}). A real-valued coherent risk functional is a distributionally robust expectation for a particular family of distributions \cite[Th. 6.6]{shapiro2009lectures}. In addition, there are connections between nested risk functionals and distributional robustness, e.g., distributionally robust performance criteria \cite{shapiro2012, bauerle2021markov} and robustness to value function approximation errors suggested by experiments \cite{kose2021risk}.
VaR and CVaR quantify risk in terms of the quantiles of a distribution, whereas mean-deviation, mean-upper-semideviation, and mean-variance quantify risk in terms of deviations from the expectation.

An appropriate approach for quantifying risk is application-dependent. For example, during a typical drive of an autonomous vehicle, penalizing the variance of the distance between a desired path and the estimated path may be appropriate, but penalizing the maximum acceleration experienced by humans during a collision (or a near-collision) may be critical to alleviate the severity of injuries. 
For another example, minimizing the CVaR of the maximum flood level may be useful for designing stormwater infrastructure \cite{chapman2021risk}. Minimizing the mean-variance of the flood level is not appropriate because the variance does not differentiate between deviations above and below the mean. (For flood management, deviations above the mean are more harmful, whereas deviations below the mean are less harmful.) The intuitive interpretation of CVaR and its strong axiomatic justification have contributed to its growing popularity for risk-averse autonomous systems in recent years. In contrast, prior to the early 2000s, exponential utility tended to be the risk functional of choice and is still popular. 

Table \ref{riskfunctionalsummarytable} summarizes Section \ref{advdis}. Subsequently, we will overview numerical methods for optimizing risk-averse autonomous systems and include historical and recent developments, with emphasis on model-based approaches. 

\begin{table}[!h]
\centering
\begin{center}
    \begin{tabular}{|p{100pt}|p{135pt}|p{45pt}|p{140pt}|}
    \hline
    {\bf {Risk functional}} & {\bf {Axiomatic aspects}} & {\bf {DP on $S$}} & {\bf {Explainability}}\\
    \hline
    Expected utility \eqref{rhophi}; $\phi(z) \neq z$ and $\phi$ is not exponential & Satisfaction of axioms depends on the utility function & No & $\phi$ is chosen to represent a decision-maker's subjective preferences.\\
    \hline
    Exponential utility \eqref{eu} & Not positively homogeneous & Yes & The parameter $\theta < 0$ represents an aversion to large variance if the approximation \eqref{myapprox} is valid.\\
    \hline
    Mean-variance \eqref{mv} & Not positively homogeneous and not monotonic if $\lambda > 0$ & No & Risk is assessed in terms of a deviation with respect to $E(Z)$.\\
    \cline{1-2}
    \hline
    Mean-deviation \eqref{md} & Coherent on the domain $L^1$ if $\lambda \in [0,\frac{1}{2}]$ & No & Same as above\\
    \cline{1-2} 
    \hline
    Mean-upper-semideviation \eqref{meanupper} & Coherent on the domain $L^\text{p}$ for $\text{p} \in [1,+\infty)$ if $\lambda \in [0,1]$ & No & Same as above\\ 
    \hline
    Value-at-risk \eqref{myvar} & Not subadditive & No & $\text{VaR}_\alpha(Z)$ is the left-side $(1-\alpha)$-quantile of the distribution of $Z$ for $\alpha \in (0,1)$. \\
    \hline
    Conditional value-at-risk \eqref{cvardefdef} & Coherent on the domain $L^1$ for $\alpha \in (0,1]$ & No & $\text{CVaR}_\alpha(Z)$ represents the expectation of $Z$ in the $\alpha$-fraction of the worst cases for $\alpha \in (0,1)$; $\text{CVaR}_1(Z) = E(Z)$. \\ 
    \hline
    Nested risk functional \eqref{gennested} & $\nu_i$ can be chosen to be coherent. & Yes & Difficult to describe intuitively in general; there are connections to distributionally robust criteria when every $\nu_i$ is coherent.\\
    \hline
    \end{tabular}
\end{center}
\caption{This table summarizes considerations related to axioms, computation, and explainability for common risk functionals. We write ``Yes'' in the ``DP on $S$'' column if, for a given MDP, a dynamic program on the state space $S$ can be formulated to optimize the risk functional of a random cumulative cost $Z$ \eqref{myZcum} under regularity conditions. Such risk functionals have computational advantages. We will present examples of dynamic programs in Section \ref{history}. Proofs of the coherence of mean-deviation, mean-upper-semideviation, and conditional value-at-risk under appropriate conditions can be found in \cite[Sec. 6.3.2]{shapiro2009lectures}.}
\label{riskfunctionalsummarytable}
\end{table}

\section{Optimizing the behaviour of an autonomous system with respect to risk-averse criteria}\label{history}
%
The problem of optimizing the behaviour of an autonomous system with safety considerations has received much research attention at least over the past fifty years.
As we have introduced, two main paradigms have emerged from this research activity, the worst-case and risk-neutral paradigms. The somewhat secondary, but rapidly expanding, risk-averse paradigm forms a bridge between the two mainstream paradigms. 
For example, the problem of minimizing the expectation of $Z \in L^1$ is equivalent to minimizing the conditional value-at-risk of $Z$ at level $\alpha = 1$ \eqref{cvardefdef}. Moreover, connections between the exponential utility functional and worst-case performance criteria for autonomous systems have been known at least for the past two to three decades \cite{coraluppi1999risk}, \cite{glover1988state}, \cite[p. 99]{lofberg2003minimax}.
%
{The literature on the worst-case paradigm is vast. While a comprehensive presentation of this paradigm is out of scope, we summarize approaches for quantifying and optimizing risk in the worst-case paradigm in Table \ref{summarytableworstcase}.} In what follows, we will present methods for evaluating and optimizing system behaviour with respect to probabilistic criteria (Section \ref{stochasticsafetyaware}) and risk functionals (Section \ref{riskawareauto}).
\begin{singlespace}
\begin{table}[h!]
\centering
\begin{center}
\begin{tabular}{ |p{370pt}|p{80pt}|} 
 \hline
{\bf Worst-case approaches for quantifying and optimizing the risk of an autonomous system} & {\bf Example papers} \\
\hline 
Risk is characterized using robust or deterministic reachable or invariant sets (tubes are also included): & \\
 \hspace{1mm} $\bullet$ Survey from 1999; covers both discrete- and continuous-time models; & \cite{blanchini1999set}\\
 \hspace{1mm} $\bullet$ Discrete-time models: & \\
 \hspace{4mm} $-${Polyhedrons, ellipsoids, and linear matrix inequalities;} & {\cite{bertsekas1971minimax, wan2003efficient, nilsson2016synthesis}}\\
 \hspace{4mm} $-$Mixed monotone systems with deterministic temporal logic specifications; & \cite{coogan2017formal}\\
 \hspace{1mm} $\bullet$ Continuous-time and hybrid models: & \\
 \hspace{4mm} $-$Sum-of-squares programming for systems with polynomial dynamics; & \cite{majumdar2014control, ahmadi2019dsos}\\
 \hspace{4mm} $-$Hamilton-Jacobi partial differential equations; & \cite{margellos2011hamilton,chen2018hamilton, mitchell2005time, fisac2015reach}\\ 
 \hspace{4mm} $-${Taylor expansions to approximate nonlinear dynamics (e.g., Flow*)}; & {\cite{chen2013flow, dutta2019reachability, ivanov2020verifying}}\\
 \hspace{4mm} $-${Boolean combinations of nonlinear constraints (e.g., dReal, dReach)}; & {\cite{eggers2008sat, gao2013dreal, kong2015dreach}}\\
  \hspace{4mm} $-${Verifying neural network controllers (e.g., Verisig, ReachNN)}. & {\cite{dutta2019reachability, ivanov2020verifying, ivanov2019verisig, huang2019reachnn}}\\
  \hline
 Risk is quantified using the optimal value of a robust or deterministic optimal control problem: & \\
   \hspace{1mm} $\bullet$ Optimizing or bounding the $H_\infty$-norm of a closed-loop transfer function; & \cite{morimoto2005robust, glover1988state, bacsar2008h} \\
  \hspace{1mm} $\bullet$ Optimizing discrete-time MDPs with respect to a maximum cost; & \cite{heger1994consideration, coraluppi1999risk} \\
 \hspace{1.3mm} $\bullet$ {Optimizing a policy subject to temporal logic constraints in a discrete-time setting}. & {\cite{raman2014model}}\\
 \hline
\end{tabular} 
\end{center}
\caption{This table summarizes approaches for quantifying and optimizing risk in the worst-case paradigm with example contributions from the communities of robust control, reinforcement learning, and formal verification.}
\label{summarytableworstcase}
\end{table}
\end{singlespace}

\subsection{Using the probability of a harmful event to quantify the risk of an autonomous system} \label{stochasticsafetyaware}
It is common to quantify the risk of a stochastic system in terms of the probability of a harmful event, as this reflects the colloquial definition for risk as the ``possibility of loss or injury'' \cite{miriamwebster}. Methods that quantify risk in this manner belong to the risk-neutral paradigm because a probability is equivalent to the expectation of an indicator function. {We summarize risk-neutral methods for quantifying the risk of an autonomous system in Table \ref{summarytableprob}. The methods involve different types of algorithms, such as Q-learning, learning from demonstrations, constraint reasoning, mathematical programming (e.g., second-order cone programming), and dynamic programming (e.g., value iteration).}
\begin{table}[h!]
\centering
\begin{center}
\begin{tabular}{ |p{395pt}|p{50pt}|} 
 \hline
{\bf Risk-neutral approaches for quantifying and optimizing the risk of an autonomous system} & {\bf Example papers} \\
\hline 
Risk is the probability of a system entering an unexplored region, entering a dangerous region, or not reaching a target: & \\
\hspace{1mm} $\bullet$ Approximation of optimal policies for constrained MDPs using Q-learning; & \cite{geibel2001reinforcement, geibel2005risk}\\
\hspace{1mm} $\bullet$ Exploration by perturbing an expert's demonstrations with noise; & \cite{garcia2012safe}\\
\hspace{1mm} $\bullet$ Dynamic programming with theoretical guarantees for hybrid systems or MDPs. & \cite{abate2008probabilistic, ding2013, yang2018dynamic, summers2010verification}
\\ \hline
Risk is the minimum expectation of a (typically) cumulative cost, where an expected-value or probabilistic constraint may be included in the problem statement: & \\
\hspace{1mm} $\bullet$ Classical approaches for minimizing the expectation of a random cumulative cost; & \cite{sutton2018reinforcement, bertsekas2004stochastic, hernandez2012discrete} \\
\hspace{1mm} $\bullet$ An expected-value constraint ensures that any state can be visited from any other state; &  \cite{moldovan2012safe}\\
\hspace{1mm} $\bullet$ Model predictive control and MDPs subject to chance constraints (a chance constraint ensures that a particular probability is within a particular interval); & \cite{haskell, schildbach2014scenario}\\
\hspace{1mm} $\bullet$ {Autonomous systems with probabilistic temporal logic (PTL)} {specifications}: & \\
\hspace{4mm} $-${Tutorials from 2007 and 2011 that describe how to assess whether or not MDPs satisfy \textcolor{white}{0000} PTL specifications}; & {\cite{kwiatkowska2007stochastic, forejt2011automated}}\\
\hspace{4mm} $-${Optimal control of hybrid systems or MDPs with PTL constraints}. & {\cite{sadigh2016safe, jha2018safe, farahani2018shrinking}}\\
\hspace{1mm} $\bullet$ {The expectation is maximized over a particular set of distributions, which may be estimated from data samples.} & {\cite{bertsimas2018data, esfahani2018data, yang2020wasserstein}}\\
\hline
\end{tabular} 
\end{center}
\caption{This table summarizes approaches for quantifying the risk of an autonomous system in the risk-neutral paradigm. A common theme is the analysis and optimization of probabilities of harmful events that may arise from a system's behaviour. We include works from the communities of reinforcement learning, formal verification, and stochastic control.}
\label{summarytableprob}
\end{table}
%

Methods have been developed to answer questions about probabilistic properties of systems with theoretical guarantees, including:
\begin{enumerate}
    \item What are the initial states from which the random state trajectory stays within a desired operating region with a sufficiently large probability? \cite{abate2008probabilistic, ding2013, yang2018dynamic}
    \item What are the initial states from which the random state trajectory reaches a target set while remaining inside a desired operating region with a sufficiently large probability? \cite{ding2013, summers2010verification}
\end{enumerate}
An approach to answer the above questions is to develop a dynamic programming algorithm using indicator functions to optimize a probability of satisfactory operation. {Next, we will describe the core machinery from \cite{abate2008probabilistic, yang2018dynamic} using the model \eqref{sysmodel}}, and we note that the machinery from \cite{ding2013, summers2010verification} is closely related. Let $K \in \mathcal{B}_S$ be a desired operating region, where we would like the random state trajectory $(X_0,X_1,\dots,X_N)$ to remain, and define the indicator function $I_{K} : S \rightarrow \{0,1\}$ by
\begin{equation}
    I_K(x) \coloneqq \begin{cases} 1, & \text{if } x \in K, \\ 0, & \text{otherwise}. \end{cases}
\end{equation}
Using the indicator function $I_K$, we define a random variable $Y$ to indicate whether or not the random state trajectory remains inside $K$:
\begin{equation}\label{myY}
 Y  \coloneqq \prod_{t=0}^{N}  I_{K}(X_t).
\end{equation}
Note that $Y$ \eqref{myY} returns either one or zero, and we prefer the larger realization because we would like the random state trajectory to stay inside $K$. Hence, an optimal control problem of interest is to maximize the expectation of $Y$:
\begin{align}
  V^*(x) & \coloneqq  \sup_{\pi \in \Pi} E_x^\pi(  Y ), \quad \quad x \in S, \label{Vstar}
\end{align}
subject to the model \eqref{sysmodel}, where $\Pi$ is the class of Markov policies. The expectation $E_x^\pi(  Y )$ is the probability that the random state trajectory stays inside $K$ when the policy is $\pi \in \Pi$ and the initial state is $x \in S$. The notation $E_x^\pi(  Y )$ symbolizes that the distribution of $Y$ \eqref{myY} depends on $\pi$ and $x$. A dynamic program provides the optimal value function $V^*$ \eqref{Vstar} in principle \cite{abate2008probabilistic}. The first step is to define $J_N \coloneqq I_K$. Then, define the functions $J_t : S \rightarrow [0,1]$ for $t = N-1,\dots,1,0$ recursively backwards in time as follows:
\begin{align}\label{18}
    J_t(x) \coloneqq \sup_{u \in C} \; I_K(x) \int_{D} J_{t+1}(f_t(x,u,w)) \; p_t(\mathrm{d}w), \quad \quad x \in S.
\end{align}
It can be shown that $J_0$ and $V^*$ are equal under regularity conditions \cite{abate2008probabilistic}. 

The above method has been extended to a \emph{distributionally robust} setting, where the disturbance distribution $p_t$ need not be known but is assumed to belong to a known class of distributions $\mathcal{D}_t$ \cite{yang2018dynamic}. Here, an optimal control problem of interest is to maximize a distributionally robust expectation of $Y$ \eqref{myY}:
\begin{align}\label{Vstarbar}
    \bar{V}^*(x) & \coloneqq  \sup_{\pi \in \Pi} \inf_{\gamma \in \Gamma} E_x^{\pi,\gamma}(  Y ), \quad \quad x \in S,
\end{align}
subject to the model \eqref{sysmodel}, where $\Gamma$ is the class of strategies of an \emph{adversary}. The adversary aims to choose a disturbance distribution $p_t \in \mathcal{D}_t$ at each time $t$ to reduce the probability of satisfactory operation. In contrast, the controller aims to provide an action $u_t \in C$ at each time $t$ to increase the probability of satisfactory operation. These competing aims are formalized by defining the optimal probability $\bar{V}^*(x)$ \eqref{Vstarbar} for a given initial state $x$ in terms of a supremum over $\Pi$ and an infimum over $\Gamma$. The latter optimization is not present in the definition of $V^*(x)$ \eqref{Vstar} because there is no adversary. Returning to the distributionally robust setting \eqref{Vstarbar}, the distribution of $Y$ depends on an adversary's strategy $\gamma$, a policy $\pi$, and an initial state $x$. The notation $E_x^{\pi,\gamma}(  Y )$ for the expectation of $Y$ in \eqref{Vstarbar} symbolizes these dependencies. $\bar{V}^*$ \eqref{Vstarbar} can be computed from a dynamic program as follows. Define the function $\bar{J}_N \coloneqq I_K$, and for $t = N-1,\dots,1,0$, define the function $\bar{J}_t$ by
\begin{align}\label{20}
    \bar{J}_t(x) \coloneqq \sup_{u \in C} \; \inf_{p_t \in \mathcal{D}_t} I_K(x) \int_{D} \bar{J}_{t+1}(f_t(x,u,w)) \; p_t(\mathrm{d}w), \quad \quad x \in S.
\end{align}
Then, $\bar{V}^*$ is equivalent to $\bar{J}_0$ under regularity conditions \cite{yang2018dynamic}. 
The difference between \eqref{18} and \eqref{20} is the optimization over disturbance distributions $p_t \in \mathcal{D}_t$ in \eqref{20}. In contrast, the disturbance distribution $p_t$ is assumed to be known exactly in  \eqref{18}.

The level sets of $V^*$ \eqref{Vstar} and $\bar{V}^*$ \eqref{Vstarbar} define \emph{probabilistic safe sets} as follows:
\begin{align}
    \mathcal{S}_{\eta} & \coloneqq \{ x \in S : V^*(x) \geq \eta \},\label{my34}\\
    \mathcal{\bar{S}}_{\eta} & \coloneqq \{ x \in S : \bar{V}^*(x) \geq \eta \}, \label{my35}
\end{align}
where $\eta \in [0,1]$ is a minimum probability threshold. $\mathcal{S}_{\eta}$ and $\mathcal{\bar{S}}_{\eta}$ are sets of initial states from which a sufficient probability of satisfactory operation is guaranteed in principle (under different assumptions about the disturbance distributions). 

{The above methods are grounded in formal proofs, which hold for fairly general stochastic systems in principle. Moreover, a user can specify a risk-aversion parameter $\eta$ that represents a minimum probability threshold and characterize the state space $S$ in terms of different probabilistic safety regions \eqref{my34}--\eqref{my35}. However, in practice, there is difficulty applying the dynamic programs \eqref{18} and \eqref{20} to systems with large state spaces or long time horizons due to memory, time, and energy requirements. In practice, the methods do not apply readily to systems that lack low-dimensional models (at least without further research).} 
\subsection{Using a risk functional to quantify the risk of an autonomous system}\label{riskawareauto}
The approaches of Section \ref{stochasticsafetyaware} and Table \ref{summarytableprob} are designed to quantify probabilities or expected costs. However, their ability to evaluate the magnitudes of harmful outcomes is limited.
Regardless of the precise meaning of safe operation, it is typically unsatisfactory for a system to operate safely on average, especially if the system interacts with humans. Instead, we usually require a system to operate safely despite rare circumstances that could disturb the system's intended behaviour. An expectation is not designed to quantify rare harmful outcomes, e.g., rare large realizations of a random cost $Z$ \eqref{myZcum}. Moreover, quantifying the probabilities of harmful events is not sufficient when the events are inevitable and their severity should be reduced as much as possible.  
Example application domains include renewable energy \cite{zakaria2020uncertainty}, water resources \cite{harremoes1988stochastic, del2015comparison}, and biology \cite{rao2002control, eling2019challenges}. In these domains, noise is ubiquitous, as suggested by the above references, and reducing the magnitudes of harmful outcomes is critical. For example, it is important to reduce the difference between energy generation and consumption, limit the volume of a stormwater overflow, and minimize the severity of an adverse medicinal side effect. The importance of evaluating and minimizing the magnitudes of rare harmful outcomes that concern autonomous systems motivates our study of methods that assess random costs using risk functionals.

{The analysis and optimization of random costs incurred by autonomous systems using risk functionals have been investigated since the 1970s.
We outline different approaches in Table \ref{summaryriskfunctionals}.} 
The type of risk functional that is chosen and whether it defines an objective criterion or a constraint depend on the desired characteristics of a cost distribution and considerations of theoretical and computational complexity and explainability. We organize our presentation of the literature according to methodology. First, we will describe methods that are based on the exponential utility functional, which has a rich history. Then, we will present additional approaches, which include methods based on nested risk functionals, reductions to risk-neutral problems, and state space augmentation.

\begin{table}[h!]
\centering
\begin{center}
\begin{tabular}{ |p{270pt}|p{175pt}|} 
 \hline
{\bf Risk-averse approaches for quantifying and optimizing the risk of an autonomous system} & {\bf Example papers} \\
\hline 
Risk is the smallest value of a risk functional of a random cost: &  \\
\hspace{1mm} $\bullet$ Expected utilities (e.g., exponential utility); & 
\cite{ coraluppi1999risk, bauerlerieder, whittle1981, haskell, glover1988state, howard1972risk, jacobson1973optimal, whittle1991risk, di1999risk, borkar2002, bielecki1999risk, cavazos2011discounted, blancas2020discounted, di2007infinite, jaskiewicz2007average, anantharam2017variational,chapman2021classical, kreps1977decisionII}; \\
\hspace{1mm} $\bullet$ Quantile-based risk functionals (e.g., CVaR); & \cite{mpctacsubmission, chapman2021risk, milleryang, bauerle2011markov, haskell, bauerle2020minimizing, chow2015risk, pflug2016time, chapman2021toward, chapmanACC}; \\
\hspace{1mm} $\bullet$ Nested risk functionals. & \cite{ruszczynski2010risk, ruszczynski2014erratum, singh2018framework, bauerle2021markov, asienkiewicz2017note}.\\ 
\hline
Risk is the smallest value of an expected random cost subject to a constraint on a risk functional of a random cost; CVaR constraints are most common. & \cite{van2015distributionally, borkar2014risk, samuelson2018safety}\\
\hline
Risk is expressed in terms of temporal logic specifications that are defined using risk functionals. & {\cite{lindemann2021stl, lindemann2021reactive, barbosa2021risk, safaoui2020control}} \\
\hline
\end{tabular} 
\end{center}
\caption{This table summarizes approaches for quantifying the risk of an autonomous system in the risk-averse paradigm. These approaches use risk functionals to analyze or optimize random costs that arise from a system's behaviour. Risk functionals can be used to define objective criteria or constraints. The most common risk functional in autonomous systems research is exponential utility. Conditional value-at-risk is the second most common risk functional, and nested risk functionals are growing in popularity.}
\label{summaryriskfunctionals}
\end{table}
%
\subsubsection{Risk-averse approaches based on exponential utility}\label{exponentialutilitymdps}
To date, the most common risk functional in the autonomous systems literature is exponential utility, which originates from classical decision theory \cite{raiffa1957games}. We summarize methods that use exponential utility in Table \ref{exputilitytable}. Initial research efforts, starting in the early 1970s, focused on linear systems with quadratic costs and Gaussian noise or MDPs with finitely many states. Subsequent work in the late 1990s to present-day includes MDPs with nonfinite state spaces, learning algorithms, and mean-field games.
\begin{table}[h!]
\centering
\begin{center}
\begin{tabular}{ |p{345pt}|p{100pt}|} 
 \hline
{\bf Problem description pertaining to exponential utility } & {\bf Example papers} \\
\hline
{Minimizing exponential utility cost functionals for autonomous systems}: & \\
\hspace{1mm} $\bullet$ {Linear systems with quadratic costs and Gaussian noise} (1973--2003); & \\ \hspace{4mm} $-$Initial solutions from the early 1970s; & \cite{jacobson1973optimal, speyer1974} \\
\hspace{4mm} $-$Solutions for partially observable systems; & \cite{whittle1981, whittle1991risk}\\ 
\hspace{4mm} $-$Connections to $\mathcal{H}_\infty$-control and minimax model predictive control; & \cite{glover1988state}, \cite[p. 99]{lofberg2003minimax} \\
\hspace{1mm} $\bullet$ MDPs with finitely many states (1972--2020); & \cite{bielecki1999risk, cavazos2011discounted, blancas2020discounted} \\
\hspace{4mm} $-$First documented solution in English in 1972; & \cite{howard1972risk} \\
\hspace{4mm} $-$Connections to worst-case performance criteria in 1999; & \cite{coraluppi1999risk}\\
\hspace{4mm} $-$Development and analysis of a Q-learning algorithm in 2002; & \cite{borkar2002}\\
\hspace{1mm} $\bullet$ MDPs that are permitted to have uncountably many states (1999--2021). & \cite{di1999risk, di2007infinite, jaskiewicz2007average, anantharam2017variational, chapman2021classical} \\
\hline
Analysis of mean-field games, where the players' risk-awareness is modelled using exponential utility functionals (1999--2020). & \cite{bacsar1999nash, moon2016linear, moon2019risk, saldi2020approximate} \\
\hline
\end{tabular} 
\end{center}
\caption{This table summarizes approaches that use exponential utility to quantify the risk of an autonomous system. The dates provide the time frames of the selected papers.}
\label{exputilitytable}
\end{table}

Next, we will present an exponential utility optimal control problem and an associated dynamic programming recursion.  
Variations of the algorithm that follows can be found in several papers, including \cite[Cor. 1, Remark 1]{bauerlerieder}, \cite[Eq. 2.3]{di1999risk}, \cite[Eq. 1.1]{asienkiewicz2017note}, and our previous work \cite[Eq. 7]{chapman2021classical}. Suppose that the stage and terminal cost functions $c_0,c_1,\dots,c_N$ in the definition of $Z$ \eqref{myZcum} are nonnegative. Consider the problem of minimizing the exponential utility of $Z$ \eqref{myZcum} at level $\theta < 0$, i.e., 
\begin{align}
    V_{\theta}^*(x) & \coloneqq \inf_{\pi \in \Pi} \textstyle\frac{-2}{\theta}\log E_x^\pi(\exp(\frac{-\theta}{2} Z)), \quad x \in S,\label{vthetastar}
\end{align}
subject to the model \eqref{sysmodel}, where $\Pi$ is the class of deterministic Markov policies.
Now, define the function 
%
$J_{N}^\theta \coloneqq c_N$ to be the terminal cost function, and for $t = N-1,\dots,1,0$, define the function $J_{t}^\theta$ in terms of $J_{t+1}^\theta$ by
\begin{align}\label{exputility}
 J_{t}^\theta(x) \coloneqq \inf_{u \in C} \left(c_t(x,u) + { \textstyle \frac{-2}{\theta}} \log \left( \int_{D} \exp\Big(\textstyle{\frac{-\theta}{2}} J_{t+1}^\theta(f_t(x,u,w))\Big) \; p_t(\mathrm{d}w) \right)  \right), \quad  x \in S.
\end{align}
One can prove that $J_{0}^\theta$ and $V_{\theta}^*$ are equivalent, and there exists a policy that is optimal for $V_{\theta}^*$, provided that regularity conditions hold \cite{bauerlerieder, chapman2021classical}. 
The definition \eqref{exputility} has structural similarities to a dynamic programming recursion for the risk-neutral problem of minimizing $E_x^\pi(Z)$ over $\Pi$ \eqref{standard2}. To see this, define $\tilde{J}_{N} \coloneqq c_N$, and then define the functions $\tilde{J}_{N-1},\dots,\tilde{J}_{1},\tilde{J}_{0}$ backwards in time by
\begin{align}\label{25}
 \tilde{J}_{t}(x) \coloneqq \inf_{u \in C} \left( c_t(x,u) + \int_{D} \tilde{J}_{t+1}(f_t(x,u,w)) \; p_t(\mathrm{d}w) \right),  \quad x \in S.
\end{align}
It is known that $\tilde{J}_{0}(x) = \inf_{\pi \in \Pi} E_x^\pi(Z)$ for every $x \in S$ under regularity conditions, e.g., see \cite[Ch. 8]{bertsekas2004stochastic} and 
\cite[Ch. 3]{hernandez2012discrete}. While the expressions for $J_{t}^\theta$ \eqref{exputility} and $\tilde{J}_{t}$ \eqref{25} are similar in appearance, they are distinct. Integrals and logarithms cannot be exchanged in general.

{There has been much research about the application of the exponential utility functional to autonomous systems. Table \ref{exputilitytable} provides example papers, and more references can be found in \cite{bauerlerieder, chapman2021classical}. However, this functional has disadvantages that deserve some attention.} It is a special case of an expected utility functional, which should reflect a user's subjective preferences. An exponential utility may not appropriately reflect the preferences of a given user. 
As we have discussed in Section \ref{advdis}, the exponential utility functional lacks coherence and need not approximate a weighted sum of the mean and variance sufficiently well. We have not seen papers that quantify the error in the approximation \eqref{myapprox} formally for an autonomous system. 
In prior work, we have demonstrated empirically that choosing a more negative value for the parameter $\theta$ does not guarantee a reduced variance \cite{kevinpaper}.

Therefore, naturally, different approaches for optimizing the behaviour of an autonomous system with respect to risk-averse criteria have been studied. Such approaches tend to be more involved in theory or computation compared to approaches that are based on exponential utility. Subsequently, we will present an approach based on nested risk functionals, which have gained popularity in recent years.
%

\subsubsection{Risk-averse optimal control using nested risk functionals}\label{methodnested}
Several approaches for optimizing nested risk functionals have been developed over the past decade, which we summarize in Table \ref{summaryriskfunctionals}. This subsection outlines an approach from 2022 \cite{bauerle2021markov} in the context of the model \eqref{sysmodel}, a finite time horizon of length $N \in \mathbb{N}$, the class of deterministic Markov policies (which we denote by $\Pi$), and the cost functions $c_0, c_1,\dots,c_N$. First, let us present some preliminaries to permit the optimal value functions of interest to be defined. Let a probability space $(\Omega,\mathcal{F},\mu)$ be given, and denote $\mathcal{Z} \coloneqq L^1(\Omega,\mathcal{F},\mu)$ for brevity. For every $t \in \mathbb{T}$, let $\nu_t : \mathcal{Z} \rightarrow \mathbb{R} \cup \{+\infty\}$ be a risk functional that is translation-equivariant. In other words, for every $a \in \mathbb{R}$ and $Z \in \mathcal{Z}$, it holds that
\begin{equation}
    \nu_t(a + Z) = a + \nu_t(Z).
\end{equation}
We will use the risk functionals $\nu_0,\nu_1,\dots,\nu_{N-1}$ to define value functions recursively. For every $\pi \in \Pi$, define the function $J^\pi_N \coloneqq c_N$ to be the terminal cost function, and for $t = N-1,\dots,1,0$, define the function $J^\pi_t$ in terms of $J^\pi_{t+1}$ by
\begin{equation}\label{37}
    J^\pi_t(x) \coloneqq  c_t(x,\pi_t(x)) + \nu_t\big( J^\pi_{t+1}(f_t(x,\pi_t(x),W_t)) \big), \quad x \in S.
\end{equation}
The quantity $J^\pi_t(x)$ represents the ``risk-to-go'' at time $t$ when the state at time $t$ is $x$ and the system operates under the policy $\pi$. In the special case when $\nu_t$ provides the expectation with respect to the distribution $p_t$ of $W_t$, i.e., when the following equality holds, 
\begin{equation}\label{my424242}
    \nu_t\big( J^\pi_{t+1}(f_t(x,\pi_t(x),W_t)) \big) = 
    \int_{D}J^\pi_{t+1}(f_t(x,\pi_t(x),w)) \; p_t(\mathrm{d}w),
\end{equation}
then $J^\pi_t(x)$ \eqref{37} is called the cost-to-go.\footnote{Here, we explain the meaning of the statement ``$\nu_t$ provides the expectation with respect to the distribution $p_t$ of $W_t$'' more formally. $(\Omega,\mathcal{F},\mu)$ is a generic probability space, and $W_t : \Omega \rightarrow D$ is $(\mathcal{F},\mathcal{B}_D)$-measurable (recall Footnote \ref{technicalfootnote}). $p_t$ is the distribution of $W_t$, which means that $p_t(\underline{D}) = \mu(\{\omega \in \Omega : W_t(\omega) \in \underline{D} \})$ for every $\underline{D} \in \mathcal{B}_D$ \cite[p. 220]{ash1972probability}. A special choice for $\nu_t$ is the expectation with respect to $\mu$, that is, $\nu_t(Z) \coloneqq \int_\Omega Z(\omega) \; \mu(\mathrm{d}\omega)$ for every $Z \in \mathcal{Z} = L^1(\Omega,\mathcal{F},\mu)$. Suppose that $\nu_t$ is the expectation with respect to $\mu$, let $g : D \rightarrow \mathbb{R} \cup \{+\infty,-\infty\}$ be Borel-measurable, and assume that $g(W_t) \in \mathcal{Z}$. Then, $\nu_t(g(W_t)) \coloneqq \int_\Omega g(W_t(\omega)) \; \mu(\mathrm{d}\omega) = \int_D g(w) \; p_t(\mathrm{d}w)$ by application of \cite[Th. 1.6.12, p. 50]{ash1972probability}. Equation \eqref{my424242} applies this change-of-measure result and assumes that $J_{t+1}^\pi(f_t(x,\pi_t(x),W_t)) \in \mathcal{Z}$ for every $x \in S$, $\pi \in \Pi$, and $t \in \mathbb{T}$. \label{footnotechange}}
The recursion \eqref{37} is well-defined under regularity conditions and specific choices for $\nu_t$, including VaR, CVaR, and risk-averse exponential utility \cite{bauerle2021markov}. In particular, one must ensure that the random variable $J^\pi_{t+1}(f_t(x,\pi_t(x),W_t))$ is in the domain of $\nu_t$ for every $x \in S$, $\pi \in \Pi$, and $t \in \mathbb{T}$. 
%
An optimal value function $J_t^*$ represents an optimal risk-to-go and is defined using $J^\pi_t$ \eqref{37} as follows. For every $t \in \mathbb{T}$, define $J_t^*$ by 
\begin{align}
    J_t^*(x) & \coloneqq \inf_{\pi \in \Pi} J^\pi_t(x), \quad x \in S, \label{Jstar}
\end{align}
subject to the model \eqref{sysmodel}, and define $J_N^*  \coloneqq c_N$. Under regularity conditions, for $t = N-1,\dots, 1,0$, $J_t^*$ satisfies the following recursion:
\begin{equation}\label{recbaurle}
    J_t^*(x) = \inf_{u \in C} \Big(c_t(x,u) + \nu_t\big(J_{t+1}^*(f_t(x,u,W_t)) \big) \Big), \quad x \in S,
\end{equation}
and there exists a policy that is optimal for $J_0^*$ \cite[Th. 4.8]{bauerle2021markov}. Moreover, observe the similarities between the risk-neutral recursion \eqref{25} and the recursion \eqref{recbaurle} when $\nu_t$ is the expectation with respect to $\mu$ (Footnote \ref{footnotechange}). 

A nested risk functional admits a dynamic program, whose iterates are defined on the state space $S$ \eqref{recbaurle}. 
This offers a computational advantage compared to approaches that we will discuss in the next two subsections (Section \ref{reducing}, Section \ref{statespaceaug}). However, a nested risk functional need not admit an intuitive interpretation. The quantity $J_0^*(x)$ \eqref{Jstar} is the optimal value of a composition of risk functionals, but this concept is challenging to relate to systems with specific safety requirements or performance goals. This issue may limit the application of nested risk functionals to the verification of autonomous system behaviour. In subsequent sections, we will introduce approaches for risk-averse optimal control that involve additional types of risk functionals.

\subsubsection{Reducing a risk-averse problem to a family of risk-neutral problems}\label{reducing}
A tactic to solve a risk-averse optimal control problem, where the risk functional is not a nested one or exponential utility,
is to reduce the problem to a family of risk-neutral problems. We will illustrate this approach using the class of extremal risk functionals.
Recall from Section \ref{extremalriskfunctionals} that an extremal risk functional $\rho_{\text{m}}$ of a random cost $Z$ takes the form
\begin{equation}\label{ext2}
    \rho_{\text{m}}(Z) = \inf_{s \in \mathbb{R}^\text{q}} E\big(g(Z,s)\big),
\end{equation}
where $g : \mathbb{R} \times \mathbb{R}^\text{q} \rightarrow \mathbb{R}$ is a convex function and $\text{q} \in \mathbb{N}$. Now, consider an optimal value function
\begin{equation}\label{Vm}
    V_{\text{m}}^*(x) \coloneqq \inf_{\pi \in \Pi} \inf_{s \in \mathbb{R}^\text{q}} E_x^\pi\big(g(Z,s)\big), \quad x \in S,
\end{equation}
subject to a stochastic system, e.g., \eqref{sysmodel}, where $\Pi$ is a class of history-dependent policies and $g$ is bounded below. 
%
%
Since we are permitted to exchange the order of infima, we can express $V_{\text{m}}^*$ as a family of risk-neutral optimal control problems as follows:
\begin{equation}\label{333}
    V_{\text{m}}^*(x) = \inf_{s \in \mathbb{R}^\text{q}} J_{\text{m}}^*(x,s), \quad x \in S,
\end{equation}
where the function $J_{\text{m}}^*$ is defined by
\begin{equation}\label{34}
    J_{\text{m}}^*(x,s) \coloneqq \inf_{\pi \in \Pi} E_x^\pi\big(g(Z,s)\big), \quad (x,s) \in S \times \mathbb{R}^\text{q}.
\end{equation}

The equations \eqref{333}--\eqref{34} provide an exact characterization of the risk-averse optimal control problem \eqref{Vm} in terms of a family of (nonstandard) risk-neutral optimal control problems. There are different algorithms to solve the standard risk-neutral problem \eqref{standard2}. For instance, algorithms based on dynamic programming, Q-learning, and linear programming are well-established \cite{sutton2018reinforcement, bertsekas2004stochastic, hernandez2012discrete}. While these algorithms are useful, they cannot apply directly to solve \eqref{34}. Additional techniques are required. In the next section, we will illustrate a common technique called state space augmentation. 
%

\subsubsection{Risk-averse optimal control using state space augmentation} \label{statespaceaug} 
We begin our discussion by first recalling the classical risk-neutral problem \eqref{standard2}. The problem of minimizing the expected cumulative cost $E_x^\pi(Z)$ subject to the model \eqref{sysmodel} can be solved by formulating a dynamic program on the state space $S$ \eqref{25}. 
A key reason is that for any Markov policy, the conditional expectation of the random variable
\begin{equation}
    Z_t \coloneqq c_N(X_N) + \sum_{i = t}^{N-1} c_i(X_i,U_i), \quad t \in \mathbb{T},
\end{equation}
given a current state $x_t \in S$, depends on $x_t$ but does not depend on the states or actions before time $t$. That is,
a current state $x_t$ can provide sufficient information about the past to choose an action for time $t$. We have seen that the problems of optimizing the exponential utility of $Z$ \eqref{myZcum} and a nested risk functional of $Z$ can also admit dynamic programs on $S$; recall \eqref{exputility} and \eqref{recbaurle}, respectively. However, problems with objectives that are defined using other risk functionals, such as CVaR, mean-variance, and expected (nonexponential) utility, do not admit dynamic programs on $S$ in general. 
%

A technique that may be useful for circumventing this issue is \emph{state space augmentation}. The key idea is to define the dynamics of an additional state to summarize enough information about the past. Then, the condensed information (rather than a complete record of the past) informs decisions for the present. The concept dates back at least to a paper from 1977, in which a ``summary space'' was proposed to study expected utility optimal control problems \cite{kreps1977decisionII}.\footnote{State space augmentation is also helpful for studying systems with partially observable states \cite[Ch. 10]{bertsekas2004stochastic}. We focus on systems with fully observable states in this survey.} Researchers have continued to apply the technique of state space augmentation to solve risk-averse problems up to present-day (Table \ref{statespaceaugtable}). The technique involves three main steps: 
\begin{enumerate}
    \item define a random additional state $Y_t$, whose realizations are in a space $\mathcal{Y}$, 
    \item define an augmented state space $S \times \mathcal{Y}$, which is the Cartesian product of the original state space $S$ and the space $\mathcal{Y}$ from the previous step, and
    \item define an algorithm that uses the random augmented state $(X_t,Y_t)$, and show that the algorithm solves the risk-averse problem of interest under regularity conditions.
\end{enumerate}
%
{Algorithms based on dynamic programming, Q-learning, and linear programming that use augmented state spaces have been developed to solve different risk-averse optimal control problems (Table \ref{statespaceaugtable}).}
An algorithm involving an augmented state space $S \times \mathcal{Y}$ typically requires more computational resources compared to an algorithm involving $S$ alone. However, the added computational effort may be worthwhile for systems with safety or performance criteria that operate under uncertainty, which has motivated a growing body of research. 

\begin{table}[h!]
\centering
\begin{center}
\begin{tabular}{ |p{145pt}|p{150pt}|p{140pt}|} 
 \hline
{\bf Type of algorithm with state space augmentation} & { Quantile-based risk functionals (e.g., CVaR)} & {Expected (nonexponential) utility risk functionals}\\
\hline
Dynamic programming or Q-learning & \cite{chapman2021risk, bauerle2011markov,bauerle2020minimizing, chow2015risk, pflug2016time, chapman2021toward, borkar2014risk} & \cite{bauerlerieder}\\
\hline
Linear programming & \cite{haskell}& \cite{haskell}\\
\hline
\end{tabular} 
\end{center}
\caption{This table presents risk-averse approaches for optimizing the behaviour of an autonomous system that apply state space augmentation, with contributions from 2011 to 2021. The linear programming category pertains to linear programming on infinite-dimensional vector spaces (not Euclidean spaces). The linear programming and dynamic programming algorithms are ``duals'' of one another, e.g., see \cite[p. ix]{hernandez2012discrete}. Among the papers above, only the paper \cite{borkar2014risk} develops a Q-learning algorithm.}
\label{statespaceaugtable}
\end{table}

{In the rest of this subsection, we will illustrate how state space augmentation applies to risk-averse autonomous systems by providing an example. We consider the problem of minimizing the CVaR of a random maximum cost subject to the model \eqref{sysmodel} from our prior work \cite{chapman2021risk}.} A random maximum cost is relevant for assessing the maximum distance between a stochastic system's trajectory and a desired operating region. 
Our method takes inspiration from Hamilton-Jacobi reachability analysis (see example references in Table \ref{summarytableworstcase}) and an approach for minimizing the CVaR of a random cumulative cost for an MDP \cite{bauerle2011markov}.\footnote{In 2021, B{\"a}uerle and Glauner generalized the approach from \cite{bauerle2011markov} to the class of \emph{spectral risk functionals} \cite{bauerle2020minimizing}. A spectral risk functional can be expressed as an integral over CVaR$_\alpha$ with respect to $\alpha$ \cite[Prop. 2.5]{bauerle2020minimizing}.} 

The problem of interest is to minimize the CVaR of a random maximum cost $\bar{Z}$:
\begin{align}
    V^*_{\alpha}(x) & \coloneqq \inf_{\pi \in \Pi} \text{CVaR}_{\alpha,x}^\pi(\bar{Z}), \quad \alpha \in (0,1], \quad x \in S,\label{27}\\
    \bar{Z} & \coloneqq \max\{c_N(X_N), c_t(X_t,U_t) : t \in \mathbb{T}\}, \label{myZbar}
\end{align}
subject to the model \eqref{sysmodel}, where $x$ is an initial state, $\alpha$ is a risk-aversion level, and $\Pi$ is a class of history-dependent policies.
We use the notation $\text{CVaR}_{\alpha,x}^\pi$ instead of $\text{CVaR}_{\alpha}$ in \eqref{27} to emphasize that the distribution of $\bar{Z}$ \eqref{myZbar} depends on $x$ and $\pi$. The stage and terminal cost functions $c_0, c_1, \dots, c_N$ are bounded below by $a \in \mathbb{R}$ and above by $b \in \mathbb{R}$ with $a \leq b$; this will allow us to specify a random additional state $Y_t$, whose realizations are in a compact interval $\mathcal{Y}$. Using the definition of CVaR \eqref{cvardefdef}, we express the optimal value $V^*_{\alpha}(x)$ \eqref{27} in terms of a family of risk-neutral problems as follows:
\begin{align}
 V^*_{\alpha}(x)
 & = \underset{s \in \mathbb{R}}{\inf} \big( s + {\textstyle\frac{1}{\alpha}}  V^s(x) \big), \label{30}\\ 
  V^s(x) & \coloneqq \inf_{\pi \in \Pi} E_x^\pi(\max\{\bar{Z}-s,0\}), \quad s \in \mathbb{R}, 
\end{align}
for every $\alpha \in (0,1]$ and $x \in S$. To derive a dynamic program to compute $V^s$, we specify $Y_1, Y_2, \dots, Y_N$ recursively by
\begin{equation}\label{46}
    Y_{t+1} = \max\{c_t(X_t,U_t), Y_t\}, \quad  t \in \mathbb{T},
\end{equation}
where the realizations of $Y_0$ are concentrated at $a$.
$Y_{t+1}$ summarizes the random maximum cost up to time $t$ because we have
\begin{equation}
    Y_{t+1} = \max\{Y_0, c_0(X_0,U_0), \dots, c_t(X_t,U_t) \}
\end{equation}
by \eqref{46} and the realizations of $Y_0$ are concentrated at a lower bound for the functions $c_0, c_1, \dots, c_N$. The realizations of $Y_t$ are in $\mathcal{Y} = [a,b]$. The random augmented state is $(X_t,Y_t)$, and the realizations of $(X_0,Y_0)$ are concentrated at $(x,a)$, where $x \in S$ is arbitrary. 

To compute $V^s$, we consider a dynamic program on $S \times \mathcal{Y}$, whose iterates are parametrized by $s \in \mathbb{R}$. Let $s \in \mathbb{R}$ be given, and define the functions $J_N^s, \dots, J_1^s, J_0^s$ recursively backwards in time as follows. 
Define the function $J_N^s$ by $J_N^s(x,y) \coloneqq \max\{\max\{c_N(x),y\} -s,0\}$ for every $(x,y) \in S \times \mathcal{Y}$, and for $t = N-1,\dots,1,0$, define the function $J_t^s$ by
\begin{equation}
    J_t^s(x,y) \coloneqq \inf_{u \in C} \int_D J_{t+1}^s(f_t(x,u,w),\max\{c_t(x,u),y\}) \; p_t(\mathrm{d}w), \quad (x,y) \in S \times \mathcal{Y}.
\end{equation}
Under regularity conditions, one can show that $J_0^s(x,a) =V^s(x)$ for every $x \in S$, and there exists a deterministic history-dependent policy that is optimal for $V^s$ \cite{chapman2021risk}. Such a policy is parametrized by $s$ and takes the form $\pi_s^* = (\pi_0^s, \pi_1^s, \dots, \pi_{N-1}^s)$, where each $\pi_t^s$ is a function from $S \times \mathcal{Y}$ to $C$ \cite{chapman2021risk}. 

Limitations of this type of policy deserve some discussion. 
Each $\pi_t^s$ is a function on the augmented state space $S \times \mathcal{Y}$ rather than just the original state space $S$, which adds computational burden. In addition, an optimal $s$ depends on a fixed initial state $x$ and risk-aversion level $\alpha$ based on \eqref{30}, implying that an optimal policy also depends on $x$ and $\alpha$.\footnote{A policy that depends on a fixed initial time and state is called a \emph{precommitment} policy, e.g., see \cite[p. 549]{bjork2014theory}.} Despite these issues, policies on augmented state spaces are of theoretical interest and have been studied in numerous risk-averse contexts \cite{bauerlerieder, bauerle2011markov,haskell, bauerle2020minimizing, chapman2021toward}.

\subsection{Scalability and adaptability challenges of risk-averse model-based approaches}
{Thus far, we have focused on presenting model-based approaches, which assume knowledge of the dynamics equations $f_t$ and disturbance distributions $p_t$ in particular. 
%
Many systems of interest operate according to well-established physical or chemical laws in practice. Thus, assuming that the dynamics equations are known need not be limiting. However, systems can involve many interacting components, e.g., robots with many joints, city-wide water networks, and several cancer drugs applied in combination, leading to high-dimensional models that are challenging to analyze and expensive to optimize.} Moreover, even when a system can be well-represented by a low-dimensional model, its intended behaviour can be affected by disturbances that are difficult to anticipate. While one can conduct experiments to estimate a family of plausible distributions for disturbances, the family can change in unexpected ways during the system's operation. In particular, assuming exact knowledge of a disturbance distribution may lead to a harmful outcome in practice, if the distribution at runtime differs significantly from the one that is assumed during planning. These issues are well-known in the autonomous systems community. 

{In the rest of the survey, we will discuss recent work and future directions that have the potential to alleviate these issues for risk-averse autonomous systems. In Section \ref{emerging}, we will present model-free methods for risk-averse optimal control. Then, in Section \ref{conclusion}, we will present future directions, which include ideas about real-time adaptation and combined model-free and model-based methods.} 

\section{Model-free methods for risk-averse optimal control}\label{emerging}
%
Algorithms that rely on simulators have the benefit of not requiring explicit mathematical models. Q-learning and temporal difference learning are classical examples of model-free algorithms.\footnote{We use the term model-free to describe methods that rely on simulation. An early exploration of (non-risk-aware) temporal difference learning was conducted by Witten in 1977 in an adaptive control setting \cite{witten1977adaptive}. The classical form of Q-learning is attributed to Watkins' research from 1989 and 1992 \cite{watkins1989learning, Watkins1992}. Additional history is provided by Sutton and Barto \cite{sutton2018reinforcement}.} Model-free algorithms require experimental platforms (in hardware or software) that can generate sufficiently many representative samples and mathematical assumptions about the exploration process. For example, to guarantee that a Q-learning algorithm converges, all state-action pairs must be simulated infinitely often.
While these requirements are suitable only for some applications, {model-free risk-averse methods have been growing in popularity since the 1990s. Table \ref{simapproach} provides example papers.} 
{In the rest of the section, we will discuss a recent work from 2021 \cite{kose2021risk}, which develops a scalable risk-averse temporal difference learning algorithm and its theoretical foundations.}

\begin{table}[h!]
\centering
\begin{center}
\begin{tabular}{ |p{365pt}|p{80pt}|} 
 \hline
{\bf Risk-averse Q-learning or temporal difference (TD) learning (1994--2021)} & {\bf Example papers}\\
\hline
$\bullet$ Q-learning with a maximum cost criterion; one of the first papers to introduce risk functionals to the machine learning community (1994);  & \cite{heger1994consideration} \\
%
$\bullet$ Q- and TD-learning algorithms that highlight unfavorable state transitions using a scalar weighting (2002); & \cite{mihatsch2002risk} \\
%
$\bullet$ Q-learning with the exponential utility functional (2002); & \cite{borkar2002} \\
%
$\bullet$ Q-learning to optimize approximately an MDP subject to a chance constraint (2001, 2005); & \cite{geibel2001reinforcement, geibel2005risk} \\
$\bullet$ Q-learning with a utility-based risk functional; the work studies connections between the algorithm and human behaviour (2014); & \cite{shen2014risk} \\
$\bullet$ Q-learning with a nested risk functional (2017, 2021); & \cite{huang2017risk, huang2020stochastic} \\
$\bullet$ Temporal difference learning with a nested risk functional (2021). & \cite{kose2021risk}\\
\hline
\end{tabular} 
\end{center}
\caption{This table summarizes risk-averse model-free approaches for autonomous systems. Most papers concern finite-state finite-action Markov decision processes. The dates provide the time frames of the selected papers.}
\label{simapproach}
\end{table}

First, we will introduce the problem setting of \cite{kose2021risk}, and then, we will present the value function of interest. The system is a finite-state finite-action MDP that operates on an infinite time horizon using a deterministic Markov policy $(\pi,\pi, \dots)$, where $\pi : S \rightarrow C$ is given. We can simulate transitions between states, but we do not know the matrix of conditional probabilities 
explicitly. For convenience, we recall some notation from Section \ref{systemmodelsec}. The state space is $S = \{1,2,\dots,\ell\}$, where $\ell \in \mathbb{N}$ is given. The scalar $\textbf{p}_{ij}(\pi(i))$ \eqref{8b} is the probability that the realization of $X_{t+1}$ is $j$, provided that the realization of $(X_t,U_t)$ is $(i,\pi(i))$. The vector $P_i^\pi \in \mathbb{R}^{1 \times \ell}$ \eqref{matrixpi} is row $i$ of the matrix of conditional probabilities $P^\pi\in \mathbb{R}^{\ell \times \ell}$. 
The goal is to estimate the long-term risk of the system under the policy $(\pi,\pi,\dots)$, which is quantified using the following value function:
\begin{equation}\label{vpi}
    v^{\pi}(i) \coloneqq \lim_{N \rightarrow \infty} \rho_{0,N}^\pi(i), \quad i \in S. 
\end{equation}
The scalar $\rho_{0,N}^\pi(i) \in \mathbb{R}$ is the value of a nested risk functional to be described. For every $t \in \mathbb{T} \cup \{N\}$, the scalar $\rho_{t,N}^\pi(i)$ is viewed as the risk of the system that is incurred from time $t$ to time $N$, provided that the realization of $X_t$ is $i$ and the system uses the policy $(\pi,\pi,\dots)$. This quantity is defined recursively in terms of a stage cost function $c: S \times C \rightarrow \mathbb{R}$ as follows. Define $\rho_{N,N}^\pi(i) \coloneqq c(i,\pi(i))$ for every $i \in S$, and then define $\rho_{t,N}^\pi$ for $t =N-1,\dots,1,0$ using a backwards recursion:
\begin{align}
    \rho_{t,N}^\pi(i) & \coloneqq c(i,\pi(i)) + \beta \cdot \zeta_i(P_i^\pi,\rho_{t+1,N}^\pi), \quad i \in S,
\end{align}
where $\beta \in (0,1)$ is a discount factor and $\zeta_i(P_i^\pi,\rho_{t+1,N}^\pi) \in \mathbb{R}$ is the value of a risk functional. The paper \cite{kose2021risk} provides a numerical example using the mean-upper-semideviation on $L^1$ \eqref{meanupper}. In the risk-neutral setting, one would choose $\zeta_i(P_i^\pi,\rho_{t+1,N}^\pi)$ to be a conditional expectation of $\rho_{t+1,N}^\pi$, i.e.,
\begin{equation}
    \zeta_i(P_i^\pi,\rho_{t+1,N}^\pi) = \sum_{j \in S} \rho_{t+1,N}^\pi(j) \; \textbf{p}_{ij}(\pi(i)).
\end{equation}
%
We will summarize the steps for estimating $v^\pi$ \eqref{vpi} from \cite{kose2021risk}, and then, we will discuss advantages and disadvantages of the proposed approach. There are three main steps to estimate $v^\pi$:
\begin{enumerate}
    \item Under regularity conditions, argue that $v^\pi$ \eqref{vpi} exists and satisfies the fixed point equation:
    \begin{equation}\label{61}
        v^\pi(i) = c(i,\pi(i)) + \beta \cdot \zeta_i(P_i^\pi,v^\pi), \quad i \in S.
    \end{equation}
    \item Assume that $v^\pi$ is well-approximated by a linear combination of functions $\varphi_j : S \rightarrow \mathbb{R}$ with $j \in \{ 1,2, \dots, M\}$ and $M \in \mathbb{N}$: 
    \begin{equation}\label{62}
    v^\pi(i) \approx \begin{bmatrix} \varphi_1(i) & \varphi_2(i) & \cdots & \varphi_M(i) \end{bmatrix}  r^{\pi} =  \varphi(i)^\top  r^{\pi}, \quad i \in S,
\end{equation}
for some vector $r^\pi \in \mathbb{R}^M$, where the symbol $\top$ denotes transpose. For convenience, define the matrix $\Phi \in \mathbb{R}^{\ell \times M}$ of features by 
\begin{equation}
    \Phi \coloneqq \begin{bmatrix} \varphi(1) & \varphi(2) &  \cdots &  \varphi(\ell) \end{bmatrix}^\top.
\end{equation}
\item Consider a realization $(i_0^\pi,i_1^\pi,\dots) \subseteq S$ of the system's trajectory under the policy $(\pi,\pi,\dots)$. As the system evolves, one applies a temporal difference algorithm to obtain a sequence $(r_0^\pi, r_1^\pi, \dots) \subseteq \mathbb{R}^M$. The vector $r_t^\pi$ for a sufficiently large $t \in \mathbb{N}$ provides an estimate for $r^\pi$. A basic version of the algorithm, which resembles gradient descent, is given by
\begin{subequations}\label{alg}
\begin{align}
    r_{t+1}^\pi & = r_t^\pi - \gamma_t \cdot g^\pi(i_t^\pi, r_t^\pi),  \quad  t = 0,1,2,\dots,
\end{align}
where $\gamma_t > 0$ is a step size, and the function $g^\pi : S \times \mathbb{R}^M \rightarrow \mathbb{R}^M$ takes the form
\begin{align}\label{64b}
    g^\pi(i,r) & = \varphi(i) \left(\varphi(i)^\top  r - c(i,\pi(i)) - \beta \cdot \hat{\zeta}_i(P_i^\pi, \Phi r) \right), \quad (i,r) \in S \times \mathbb{R}^M.
\end{align}
\end{subequations}
The scalar $\hat{\zeta}_i(P_i^\pi, \Phi r)$ is an estimate for $\zeta_i(P_i^\pi, \Phi r)$, which is obtained by sampling transitions from state $i$. An estimate  for $\zeta_i(P_i^\pi, \Phi r)$ is needed because the vector of conditional probabilities $P_i^\pi$ is not known. $g^\pi(i,r)$ \eqref{64b} approximates the gradient of a mean-squared error with respect to $r$.\footnote{The error is the difference between a linear approximation for $v^\pi$ \eqref{62} and an estimate for the right side of the fixed point equation \eqref{61}:
\begin{equation}\label{myerror}
     \text{error}_i^\pi(r) \coloneqq \varphi(i)^\top  r - c(i,\pi(i)) - \beta \cdot \hat{\zeta}_i(P_i^\pi,\Phi r).
\end{equation}
$g^\pi(i,r)$ \eqref{64b} equals the gradient of $\frac{1}{2}(\text{error}_i^\pi(r) )^2$ with respect to $r$, provided that the term $\hat{\zeta}_i(P_i^\pi,\Phi r)$ in \eqref{myerror} is treated as being constant in $r$.} The sequence $(r_0^\pi, r_1^\pi, \dots)$ is studied formally in \cite{kose2021risk} under appropriate conditions.
\end{enumerate}

%
{The above method does not require all  state-action pairs to be visited infinitely often; in contrast, Q-learning algorithms have this exploration requirement. Moreover, the method scales to large finite state spaces due to the assumed linear approximation for $v^\pi$ \eqref{62}. A linear system example with $\ell$ on the order of $10^{427}$ states has been implemented on a four-core computer \cite{kose2021risk}. We are not aware of another risk-averse method that scales so well, and it is likely that other researchers will develop related frameworks to gain the scalability benefit. 
However, the assumption that $v^\pi$ is well-approximated by a linear combination of features \eqref{62} lacks theoretical justification. It is not known how to select the features a priori 
to guarantee a sufficient degree of accuracy. 
Quantifying the errors induced by the feature approximation, i.e., $v^{\pi}(i) - \varphi(i)^\top  r_t^\pi$ for some $t \in \mathbb{N}$ and $i \in S$, is an open question. An assessment of these errors is valuable for any system and is especially important for systems with safety concerns.} 
\section{Conclusions and future directions}\label{conclusion}
Model-free and model-based approaches have advantages and disadvantages, which have particular relevance for risk-averse autonomous systems. Of course, simulations in software and experiments in hardware have limited ability to represent the nuances of the real world. 
Using simulations or experiments to anticipate and circumvent \emph{rare} harmful outcomes, which concern risk-averse systems, is difficult. This task may require the selective generation of samples, e.g., importance sampling \cite{hanna2021importance}. Otherwise, the task may require large numbers of samples, which is not feasible for every application.
For example, if the task is to design a treatment strategy for a leukemia patient, collecting many identically distributed blood samples is not practical due to inconvenience, discomfort, and financial limitations. While model-based approaches often enjoy desirable theoretical properties in principle, their practical use requires sufficiently representative disturbance distributions and low-dimensional dynamics equations, which may not be available.

{Safe learning is a growing research area that builds on ideas from adaptive control theory \cite{sastry1989adaptive} and reinforcement learning to develop combined model-free and model-based methods for autonomous systems. Let us briefly mention two early works in safe RL. In 1996, Schneider examined an application of dynamic programming in which the assumed model is reidentified continually using up-to-date observations to promote safer operation \cite{schneider1997exploiting}. Six years later, Perkins and Barto applied Lyapunov stability theory to design policies with safety and performance guarantees \cite{perkins2002lyapunov}. In recent years, safe learning has focused on improving the estimates of models and reachable (or invariant) sets using data samples and Gaussian processes. Surveys from 2020 and 2022 are provided by \cite{hewing2020learning} and \cite{Brunke}, respectively. Future work that merges machinery from safe learning and risk-averse optimal control is needed to permit real-time adaptation of risk-averse autonomous systems to changing environments. We are not aware of any research that uses risk functionals and data samples to improve the estimates of models or reachable sets in real time. However, this research direction is likely to be fruitful, given experimental evidence that some risk functionals can offer protection against modelling errors \cite{kose2021risk}.}

{Typically, a risk functional and a risk-aversion parameter are fixed a priori, and then, an algorithm is designed to optimize a random cost that is evaluated according to these choices. It would be worthwhile to update the value of the risk-aversion parameter iteratively, as new data is acquired in real time. That is, concepts from adaptive control and safe learning could be applied to learn the risk-aversion parameter's value and an action for time $t$ based on the current estimate of the model's accuracy.} 

Below, we outline additional areas for future research:
\begin{itemize}
    \item The field requires more knowledge about how to design experiments so that the experiments provide data samples with desired statistical properties at runtime.
    \item Studies about risk-averse model-free methods from a nonasymptotic viewpoint are needed. This is because finite time horizons and finitely many samples are used in practice. It is notable that a paper from 2021 \cite{huang2020stochastic} offers a nonasymptotic analysis of a risk-averse Q-learning algorithm. 
    \item There is limited understanding about which risk functional may be more useful for a given application. To improve this understanding, the field requires additional empirical studies that compare the performance of different systems, where performance is quantified using different risk functionals. For instance, in previous experimental work, we discovered that choosing a more negative value for $\theta$ when optimizing exponential utility \eqref{vthetastar} can increase both the mean and variance, suggesting caution when using exponential utility for optimal control in practice \cite{kevinpaper}.
    \item New theoretical risk-averse optimal control methods that include both model-free and model-based aspects will likely yield positive impact.
\end{itemize}


Naturally, an appropriate blend of model-free and model-based approaches is application-specific. 
For example, in online shopping or ride-sharing, data is plentiful and is likely suitable for predicting consumer behaviour. However, in medical or environmental applications that involve rare high-consequence situations, simulators or representative data sets may not be available. In this case, using simulations or data to drive decision-making would not be suitable. Instead, it would be suitable to use a mixture of tools and knowledge. For instance, to help improve cancer outcomes, work with oncologists and use patient data as well as biological and chemical models. To support urban water environmental health, work with water specialists and city planners and use precipitation data, weather forecasts, and hydrologic models. 

Risk-averse autonomous systems research currently emphasizes applications related to finance, robots, and vehicles (ground or aerial). While these applications are important, diversity of applications is also important. Domains that deserve further emphasis in risk-averse systems research include renewable energy \cite{zakaria2020uncertainty}, water resources \cite{harremoes1988stochastic, del2015comparison}, and biology \cite{rao2002control, eling2019challenges}. These domains require systems to operate effectively despite uncertainties, and the mitigation of harmful consequences in these domains is paramount. We are hopeful that the technology of risk-averse autonomous systems will be developed and applied broadly to enhance human and environmental welfare. 

\section*{Declaration of competing interest}
The authors declare that they have no known competing financial interests or personal relationships that could have appeared to influence the work reported in this paper.

\section*{Acknowledgements}
{The authors are grateful for discussions with Kevin M. Smith and Yuxi Han and for constructive feedback by three anonymous reviewers. The authors appreciate conversations with Peter Caines, who suggested recent research on risk-averse mean-field games.} The authors are thankful for comprehensive presentations by A. Shapiro, D. Dentcheva, and A. Ruszczy{\'n}ski \cite{shapiro2009lectures} and N. B{\"a}uerle and U. Rieder \cite{bauerlerieder}, which provided a foundation for Section \ref{examplesriskfunctionals}. In addition, the authors thank J. Kisiala \cite{kisi} for offering intuitive introductions to CVaR, VaR, and coherent risk functionals in general. 
Y. Wang is supported in part by the Natural Sciences and Engineering Research Council of Canada (NSERC) Discovery Grants Program [RGPIN-2022-04140]. Cette recherche a \'{e}t\'{e} financée par le Conseil de Recherches en Sciences Naturelles et en G\'{e}nie du Canada (CRSNG) [RGPIN-2022-04140]. This research is supported in part by the Edward S. Rogers Sr. Department of Electrical and Computer Engineering, University of Toronto.


\bibliography{mybibfile}

\end{singlespace}
\end{document}